\documentclass[sigconf]{acmart}

\AtBeginDocument{%
  }

\copyrightyear{2024} 
\acmYear{2024} 
\setcopyright{acmlicensed}\acmConference[WWW '24]{Proceedings of the ACM
 Web Conference 2024}{May 13--17, 2024}{Singapore, Singapore}
 \acmBooktitle{Proceedings of the ACM Web Conference 2024 (WWW '24), May
 13--17, 2024, Singapore, Singapore}
 \acmDOI{10.1145/3589334.3645472}
 \acmISBN{979-8-4007-0171-9/24/05}
\usepackage[utf8]{inputenc} 
\usepackage[T1]{fontenc}    
\usepackage{hyperref}       
\usepackage{url}            
\usepackage{booktabs}       
\usepackage{amsfonts}       
\usepackage{nicefrac}       
\usepackage{microtype}      
\usepackage{xcolor}         
\usepackage{subfigure}
\usepackage{makecell}
\usepackage{listings}
\usepackage{amsmath}
\usepackage{multirow}
\usepackage{graphicx}
\usepackage[shortlabels]{enumitem}
\usepackage[ruled,linesnumbered]{algorithm2e}

\begin{document}

\title{LARA: A Light and Anti-overfitting Retraining Approach for Unsupervised Time Series Anomaly Detection}

\author{Feiyi Chen}
\affiliation{%
  \institution{Zhejiang University}
  \institution{Alibaba Group}
  \city{Hangzhou}
  \country{China}
}
\email{chenfeiyi@zju.edu.cn}

\author{Zhen Qin}
\affiliation{%
  \institution{Zhejiang University}
  \city{Hangzhou}
  \country{China}}
\email{zhenqin@zju.edu.cn}

\author{Mengchu Zhou}
\affiliation{%
  \institution{Zhejiang Gongshang University}
  \city{Hangzhou}
  \country{China}}
\email{mengchu@gmail.com}

\author{Yingying Zhang}
\affiliation{
  \institution{Alibaba Group}
  \city{Hangzhou}
  \country{China}
}
\email{congrong.zyy@alibaba-inc.com}

\author{Shuiguang Deng}
\authornote{Corresponding Author.}
\affiliation{%
  \institution{Zhejiang University}
  \city{Hangzhou}
  \country{China}
}
\email{dengsg@zju.edu.cn}

\author{Lunting Fan}
\affiliation{
  \institution{Alibaba Group}
  \city{Hangzhou}
  \country{China}
}
\email{lunting.fan@taobao.com}

\author{Guansong Pang}
\affiliation{
  \institution{Singapore Management University}
  \country{Sinapore}
}
\email{gspang@smu.edu.sg}

\author{Qingsong Wen}
\affiliation{
  \institution{Squirrel AI}
  \city{Bellevue}
  \state{WA}
  \country{USA}
}
\email{qingsongedu@gmail.com}

\renewcommand{\shortauthors}{Chen F, Qin Z, Zhou M, Zhang Y, Deng S, Fan L, Pang G, Wen Q.}

\begin{abstract}
  Most of current anomaly detection models assume that the normal pattern remains the same all the time. However, the normal patterns of 
  web services can change dramatically and frequently over time. The model trained on old-distribution data becomes outdated and ineffective after such changes. Retraining the whole model 
  whenever the pattern is changed is computationally expensive. 
  Further, at the beginning of normal pattern changes, there is not enough observation data from the new distribution. Retraining a large neural network model with limited data is vulnerable to overfitting. Thus, we propose a \textbf{L}ight
  \textbf{A}nti-overfitting \textbf{R}etraining \textbf{A}pproach (LARA) based on deep variational auto-encoders for time series anomaly detection. 
  In LARA we make the following three 
  major contributions: 1) the retraining process is designed as a convex problem such that overfitting is prevented and the retraining process can converge fast;
  2) a novel ruminate block is introduced, which can leverage the historical data without the need to store them; 3) we mathematically and experimentally prove that when fine-tuning the latent vector and reconstructed data, the linear formations can achieve the least adjusting errors between the ground truths and the fine-tuned ones.
  Moreover, we have performed many experiments to verify that retraining LARA with even 
  a limited amount of data from new distribution can achieve competitive performance in comparison with the state-of-the-art anomaly detection models trained with sufficient data. Besides, we verify its light computational overhead.
\end{abstract}



\received{13 May 2024}
\received[revised]{16 February 2024}
\received[accepted]{22 January 2024}

\maketitle

\section{Introduction}
Web services have experienced substantial growth and development in recent years \cite{luo2016effective,yuan2020energy,zhang2023predicting}. In large cloud centers, millions of web services operate simultaneously, posing significant challenges for service maintenance, latency detection, and bug identification. Traditional manual maintenance methods struggle to cope with the real-time management of such a vast number of services. Consequently, anomaly detection methods for web services have garnered widespread attention, as they can significantly enhance the resilience of web services by identifying potential risks from key performance indicators (KPIs).
Despite significant progress in anomaly detection, the high dynamicity of web services continues to present challenges for efficient and accurate anomaly detection. Large web service cloud centers undergo thousands of software updates every day \cite{ma2021jump}, which outdates the original model recurrently. Currently, web service providers rely on periodic model retraining, incurring significant computational expense. Furthermore, during the initial stages of a web service update, newly observed KPIs are scarce, making it impractical to support large neural network retraining without causing model overfitting. The time required to collect a sufficient number of KPIs can sometimes extend to tens of days \cite{ma2021jump}. During this period, the outdated models underperform, and the updated models are not yet ready, rendering web services vulnerable to intrusions and bugs. Consequently, there is a pressing need for a data-efficient and lightweight retraining method.

The existing methods to deal with this problem can be roughly divided into three categories: signal-processing-based, transfer-learning-based, and few-shot-learning-based. Among them, the first one suffers from heavy 
inference time overhead and struggles to cope with the high traffic peaks of web services in real time. 
The second one \cite{kumagai2019transfer} does not consider the chronological order of multiple historical distributions (i.e. the closer historical distribution generally contains more useful knowledge for the newly observed distribution, compared with farther ones). The last one \cite{wu2021learning} needs to store lots of data from historical distributions and also ignores the chronological orders. 

\begin{figure}[t]
  \centering
  \subfigure[The 
  reconstructed data samples]{
      \includegraphics[width=0.48\linewidth]{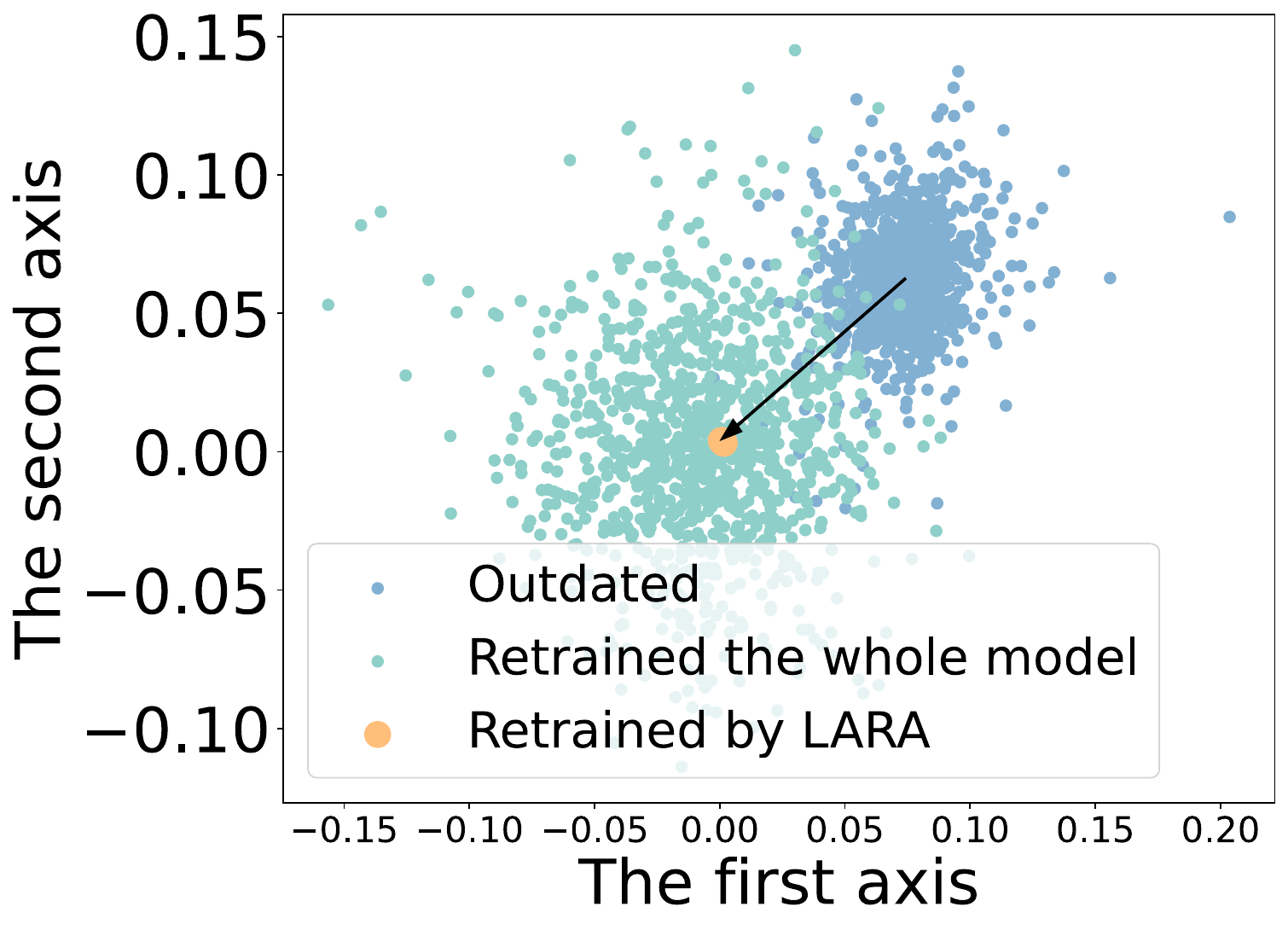}
      \label{fig:xvisual}
  }
  \hfill
  \subfigure[The 
  latent vectors]{
      \includegraphics[width=0.455\linewidth]{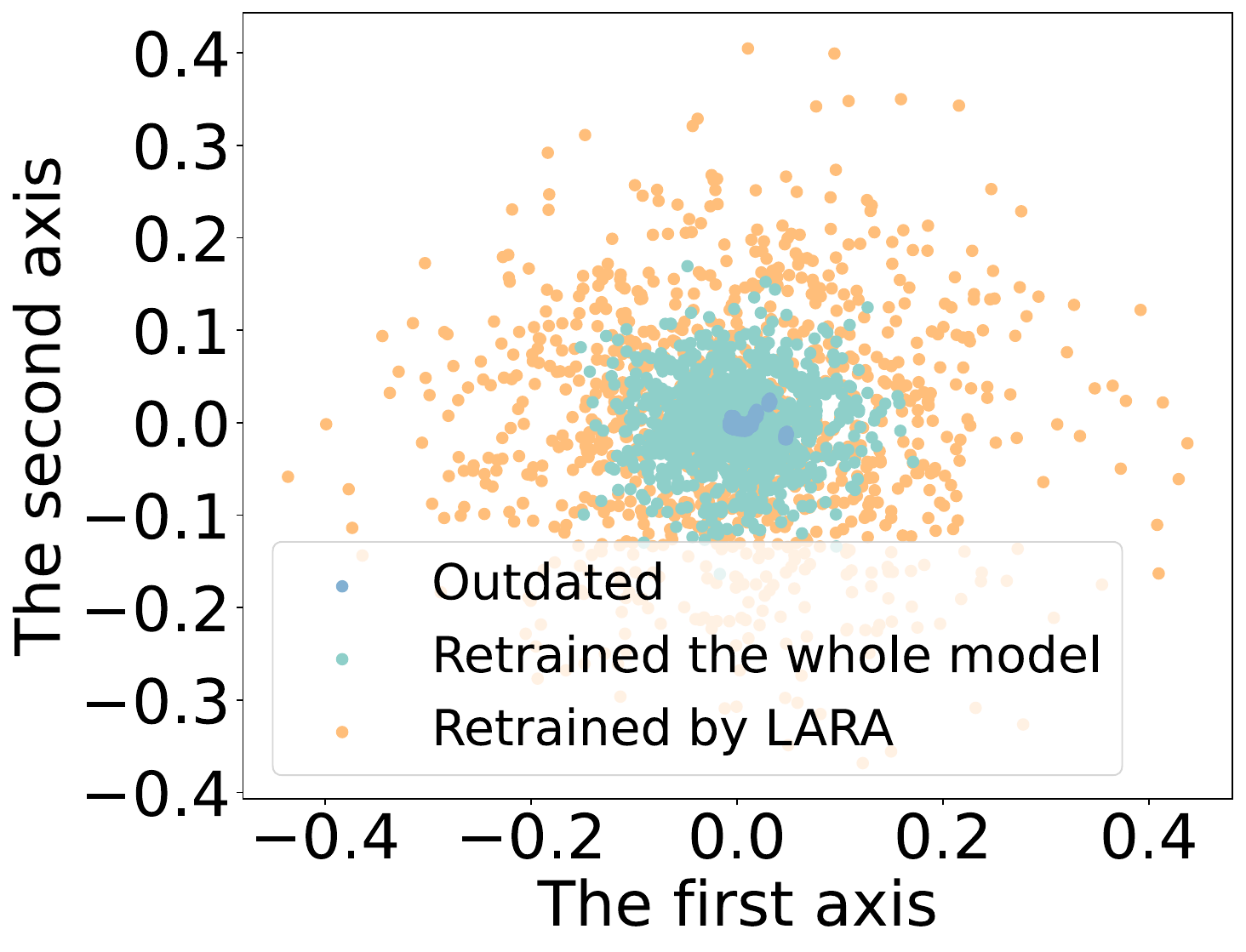}
      \label{fig:zvisual}
  }
  \vspace{-3mm}
  \caption{
  LARA vs. the other two approaches. The figures show 
  (a) reconstructed data samples and (b) latent vectors output by three different approaches: the model trained on historical data only (i.e. outdated model), the model retrained by the whole dataset, and the model retrained by LARA.}\label{fig:motivation}
\end{figure}

To overcome these drawbacks, we propose a Light and Anti-overfitting Retraining Approach (LARA) for deep Variational Auto-Encoder-based unsupervised anomaly detection models (VAEs), considering the VAEs are one of the most popular unsupervised anomaly detection methods. The VAE-based methods learn a latent vector for each input data sample and use the latent vector to regenerate it. The main idea 
of LARA is to fine-tune the latent vectors with historical and newly observed data without storing the historical data and adapt the reconstructed data samples to the new distribution. 
This enables LARA to adapt the reconstructed data samples to the center of the new distribution (Fig.\ref{fig:xvisual}) and loses the boundary of latent vectors moderately according to the historical and newly observed data (Fig.\ref{fig:zvisual}) to enhance the ability of VAEs to deal with unmet distributions. LARA achieves this via three prominent components: ruminate block, adjusting functions $M_z$ and $M_x$, and loss function design. 

Ruminate block\footnote{The ruminate block is named after the rumination of cows, which chews the past-fed data and extracts general knowledge.} leverages the historical data and newly-observed data to guide the fine-tuning of latent vectors, without storing the historical data. Its main idea is that the model trained on historical distributions is an abstraction of their data. Thus, the ruminate block can restore the historical data from the old model and use them with the newly observed data to guide the fine-tuning of the latent vector.
There are three advantages of using the ruminate block: 1) it saves storage space as there is no need to store data from historical distributions; 2) it chooses the historical data similar to the new distribution to restore, which would contain more useful knowledge than the others; and 3) with the guidance of the ruminate block, the latent vector generator inherits from the old model and is fine-tuned recurrently, which takes the chronological order of historical distributions into account.

The adjusting functions $M_z$ and $M_x$ are devised to adapt the latent vector and the reconstructed data sample to approximate the latent vector recommended by the ruminate block and the newly observed data, respectively. We mathematically prove that linear formations can achieve the least gap between the adjusted ones and ground truth. It is interesting that the adjusting formations with the least error are amazingly simple and cost light computational and memory overhead.
Furthermore, we propose a principle of loss function design for the adjusting formations, which ensures the convexity of the adjusting process. It is proven that the convexity is only related to the adjusting functions $M_z$ and $M_x$ without bothering the model structure, which makes loss function design much easier.
The convexity guarantees the $O(\frac{1}{k})$ converging rate (where $k$ denotes the number of iteration steps) and a global unique optimal point which helps avoid overfitting, since overfitting is caused by the suboptimal-point-trapping. 

Accordingly, this work makes the following novel and unique contributions to the field of anomaly detection:
\begin{enumerate}[itemindent=1em, listparindent=2em, leftmargin=0em]
  \item[1) ] We propose a novel retraining approach called LARA, which is designed as a convex problem. This guarantees a quick converging rate and prevents overfitting. 
  \item[2) ] We propose a ruminate block to restore historical data from the old model, which enables the model to leverage historical data without storing them and provides guidance to the fine-tuning of latent vectors of VAEs.
  \item[3) ] We mathematically and experimentally prove that the linear adjusting formations of the latent vector and reconstructed data samples can achieve the least adjusting error. These adjusting formations are simple and require only little time and memory overhead.
\end{enumerate}
In addition, we conduct extensive experiments on four real-world datasets with evolving normal patterns to show that LARA can achieve the best F1 score with limited new samples only, compared with the state-of-the-art (SOTA) methods. Moreover, it is also verified that LARA requires little memory and time overhead for retraining. Furthermore, we substitute $M_z$ and $M_x$ with other nonlinear formations and empirically prove the superiority of our linear formation over the nonlinear ones.

\begin{table*}[tbh]
  \centering
  \caption{\label{symbols}The definition of symbols used in this paper.}
  \vspace{-3mm}
  \begin{tabular}{l|l|l|l}
    \hline
    Symbol            & Meaning                                                & Symbol               & Meaning                                                   \\ \hline
    $D_i$             & The $i$th distribution                                 & $V_i$                & The model for $D_i$                                       \\
    $X_i$             & The data samples for $D_i$                                  & 
    $Z_{i,k}$         & The latent vectors of $X_i$ obtained by $V_k$          \\
    $\tilde{X}_{i,k}$ & The reconstructed data of $X_i$ obtained by $V_k$      & 
    $\tilde{Z}_i$     & The latent vector of $X_i$ estimated by ruminate block \\
    $M_x$             & The adjusting function of reconstructed data           & $M_z$                & The reconstructed data of latent vector                   \\
    $\mathcal{P}_x$   & The trainable parameters of $M_x$                      & $\mathcal{P}_z$      & The trainable parameters of $M_z$                         \\ \hline
    \end{tabular}%
  \end{table*}


\begin{figure*}[tbhp]
  \centering 
  \includegraphics[width=0.7\linewidth]{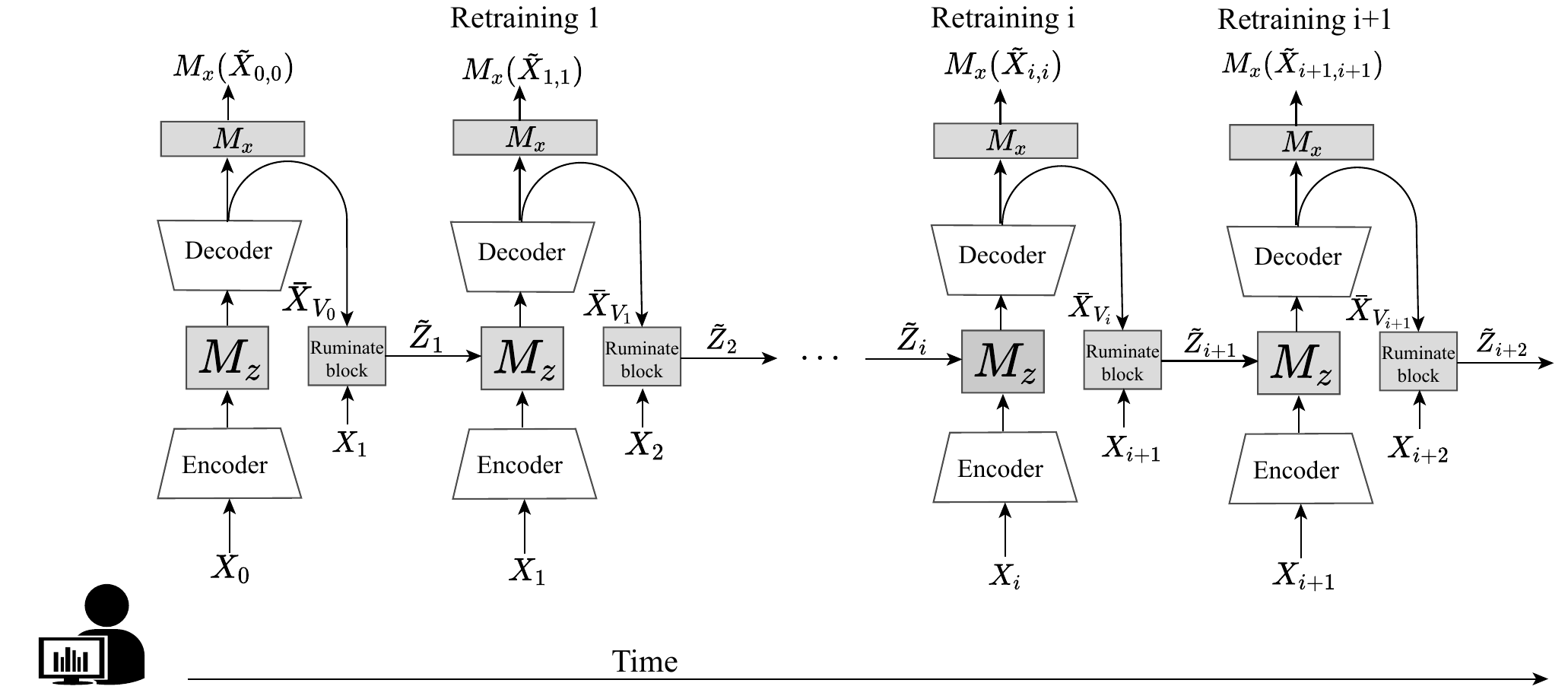} \vspace{-3mm}
  \caption{Overview of LARA. When there is a new distribution shift, LARA retrieves historical data from the latest model and uses them with a few newly observed data sample to estimate the latent vector for each new sample by the ruminate block. Then, LARA uses two adjusting functions -- $M_z$ and $M_x$ -- to adapt the latent vector to the estimated one by the ruminate block, and adapt the reconstructed sample yielded by the latest model to the sample from the new distribution.}
  \label{fig:modelArch0}
\end{figure*}

 \section{Proposed method}
\textbf{Preliminary.}
VAE-based methods are among the most popular anomaly detection models, such as omniAnomaly \cite{su2019robust}, ProS \cite{kumagai2019transfer} and Deep Variational Graph Convolutional Recurrent Network (DVGCRN) \cite{chen2022deep}. These methods are usually divided into two parts: encoder and decoder. An encoder is designed to learn the posterior distribution $p(z|x)$, where $x$ is observed data, and $z$ is a latent vector \cite{DBLP:journals/corr/KingmaW13}. A decoder is designed to learn the distribution $p(x|z)$. 
Please refer to Tab. ~\ref{symbols} for the definitions of symbols used in this paper.

 \subsection{Overview}
 The overview of LARA is given in Fig. \ref{fig:modelArch0}. On the whole, LARA leverages the data from historical distribution and newly-observed data to adjust the latent vector of VAEs and only uses the newly-observed data to adapt the reconstructed data sample. Whenever a retraining process is triggered, LARA first uses its ruminate block to restore historically distributed data from the latest model. Then, the ruminate block uses the restored data and newly-observed data to estimate the latent vector for each newly-observed data. Then, LARA employs a latent vector adjusting function $M_z$ to map the latent vector generated by the encoder to the one estimated by the ruminate block. To further adapt the model, LARA also applies an adjusting function $M_x$ to adjust the reconstructed data sample. After that, LARA uses a loss function to fit the trainable parameters in $M_x$ and $M_z$, which guarantees the convexity of the adjusting process.  

For each newly-observed data sample, the ruminate block retrieves $n$ historical data samples that are similar to the newly-observed data from the latest model. Then, the ruminate block estimates the latent vector for each newly-observed data by Online Bayesian Learning \cite{minka2013expectation}, where the restored data and the newly-observed data are regarded as the historical data and the present data in Online Bayesian Learning respectively. The details are given in Sec.  \ref{reminateSec}. 

 The adjusting functions $M_z$ and $M_x$ are mathematically proven to achieve the least adjusting error. Besides, there is an interesting finding that the best formations of $M_z$ and $M_x$ are linear, which are surprisingly simple and require low retraining overhead. The formulation of the adjusting error and the mathematical proof are given in Sec.  \ref{adjustFunc}. 
 
The simple formations of $M_z$ and $M_x$ make it possible to skillfully design a convex loss function for trainable parameters in $M_z$ and $M_x$. The convexity not only prevents the overfitting problem, as there is a unique global optimal point for the convex problem, but also guarantees a fast convergence rate for the retraining process, which contributes to the low time overhead. However, in general, designing such a loss function requires sophisticated techniques and knowledge of convex optimization. Thus, we propose the principles of designing such a convex loss function in Sec.  \ref{lossFuncSec} to make the designing process easy and convenient. It is found in Sec.  \ref{lossFuncSec} that the convexity is not related to the model structure but only related to the design of the loss function, which makes the designing process much easier since there is no need to consider the model structure. 
Any formations that satisfy the principles can guarantee convexity. According to Occam's razor principle \cite{walsh1979occam}, LARA chooses one of the simplest as its loss function.

The following subsections illustrate each module, assuming that the retraining process is triggered by newly-observed distribution $D_{i+1}$ and the latest model is $V_i$.

 
 \subsection{Ruminate block}
 \label{reminateSec}
 For each newly-observed sample $X_{i+1} $, the ruminate block firstly obtains the conditional distribution $p(\tilde{X}_{i+1,i} |Z_{i+1,i} )$ from $V_i$. Then, it generates $n$ data samples from the distribution as the historical data. There are two reasons to restore the historical data in this way: 1) the latest model $V_i$ contains the knowledge of historical distribution and its reconstructed data represents the model's understanding of the historical normal patterns; and 2) the historical data is reconstructed from the newly observed data, which selectively reconstructs data similar to the new one among all the historical distributions. 

 Furthermore, the ruminate block uses the restored historical data $\bar{X}_{i}$ and newly-observed data $X_{i+1}$ to estimate the latent vector for each newly-observed data. Inspired by the Online Bayesian Learning \cite{minka2013expectation}, the expectation and variance of the estimated latent vector, when given a specific $X_{i+1} $, is shown in the following: 
\begin{equation}
  \label{expectation1}
  \begin{matrix}
    \mathbb{E}(\tilde{Z}_{i+1} |X_{i+1} )
 =  \frac{\mathbb{E}_{z\sim p(z)}[p(\tilde{X}_{i+1} |z)p(\bar{X}_i|Z_{i+1,i})z]}{\mathbb{E}_{z\sim p(z)}[p(\tilde{X}_{i+1} |z)p(\bar{X}_i|Z_{i+1,i})]},
  \end{matrix}
\end{equation}
\begin{equation}
 \label{variance1}
 \operatorname{Var}(\tilde{Z}_{i+1} |X_{i+1} )=\mathbb{E}_z^T\mathbb{E}_z - \mathbb{E}_{z^Tz}.
\end{equation}
where $\mathbb{E}_z=\mathbb{E}(\tilde{Z}_{i+1} ^T\tilde{Z}_{i+1} |X_{i+1} )$, $p(z)$ follows the normal distribution, and $\mathbb{E}_{z^Tz}= \frac{\mathbb{E}_{z\sim p(z)}[p(\tilde{X}_{i+1,i} |z) p(\bar{X}_{i}|Z_{i+1}) z^Tz]}{\mathbb{E}_{z\sim p(z)}[p(\tilde{X}_{i+1,i} |z) p(\bar{X}_{i}|Z_{i+1})]}$.
The proof is given in Appendix \ref{proofTh3}. The expectations in Eq.\ref{expectation1}-Eq.\ref{variance1} are computed by the Monte Carlo Sampling \cite{metropolis1949monte} and the conditional distributions are given by the decoder of $V_i$.

We provide an intuitive explanation of the Eq.\ref{expectation1}-Eq.\ref{variance1} in the following. We take $\mathbb{E}(\tilde{Z}_{i+1} |X_{i+1} )$ as an example to illustrate it and the variance can be understood in a similar way. When using Monte Carlo Sampling to compute it, the expectation can be transformed into the following:
 \begin{equation}
   \label{expectationMonte}
   \begin{matrix}
    \mathbb{E}(\tilde{Z}_{i+1} |X_{i+1} )
   =\sum_{s=1}^N  \frac{\alpha_s}{\sum_{k=1}^N \alpha_k} z_s .
   \end{matrix}
 \end{equation}
 where $\alpha_s=p(\tilde{X}_{i+1,i} |z_s)p(\bar{X}_{i}|Z_{i+1,i})$ and $z_s$ is the $s$th sampling of $z$ from $p(z)$.
 
It is now obvious that $\mathbb{E}(\tilde{Z}_{i+1} |X_{i+1} )$ is a weighted summation of different $z_s$ throughout the distribution $p(z)$. Next, we look into the value of weight $\alpha_s$ to figure out under which conditions it assumes higher or lower values. It depends on both the distributions of $p(\tilde{X}_{i+1,i} |z_s)$ and $p(\bar{X}_{i}|Z_{i+1,i})$. The greater the reconstructed likelihoods of $\tilde{X}_{i+1,i}$ and $\bar{X}_{i}$ are, the greater the weight is. Thus, the estimation of the latent vector is close to the value with a high reconstructed likelihood of $\tilde{X}_{i+1,i}$ and $\bar{X}_{i}$. As shown in Fig. \ref{fig:zvisual}, this estimation looses the boundary of the latent vector, as it not only considers the reconstructed likelihood of newly observed data, but also considers the one for historical data, which contributes to dealing with the unseen distributions.

 \subsection{Functions $M_x$ and $M_z$}
 \label{adjustFunc}
In LARA, we propose an adjusting function $M_z$ to make $M_z(Z_{i+1,i})$ approximate to $Z_{i+1,i+1}$, which is estimated by $\tilde{Z}_{i+1}$. Considering that each distribution has its specialized features, we also use an adjusting function $M_x$ to make $M_x(\tilde{X}_{i+1,i})$ approximate to $X_{i+1}$. 
 
 This subsection answers two questions: 1) what formations of $M_z$ and $M_x$ are the best; and 2) could we ensure accuracy with low retraining overhead? We first give the formulation of the adjusting errors of $M_z$ and $M_x$. Then, we explore and prove the formations of $M_z$ and $M_x$ with the lowest adjusting error. We surprisingly find that the best formations are simple and require little overhead. Following \cite{DBLP:journals/corr/KingmaW13}, we make the below assumption.\\
 \textbf{Assumption 1.} $p(\tilde{X}_{i,i}|Z_{i,i})$ and $p(Z_{i,i}|X_{i})$ follow Gaussian distribution.\\
 \textbf{Assumption 2.} $p(Z_{i+1,i},Z_{i+1,i+1})$ and $p(\tilde{X}_{i+1,i},\tilde{X}_{i+1,i+1}|Z_{i+1,i})$ follow Gaussian joint distributions. \\
 \textbf{Quantifying the adjusting errors of $M_z$ and $M_x$.} We formulate the mapping errors $\mathfrak{E}_z$ and $\mathfrak{E}_x$ for $M_z$ and $M_x$ in the following:
 \begin{equation}
   \label{ErrDef1}
   \mathfrak{E}_z=\int   \mathbb{E}[(M_z(Z_{i+1,i})-Z_{i+1,i+1})^2|X_{i+1}] f(X_{i+1}) dx,
 \end{equation}
 \begin{equation}
   \label{ErrDef2}
   \mathfrak{E}_x= \mathbb{E}[(M_x(\tilde{X}_{i+1,i})-\tilde{X}_{i+1,i+1})^2|Z_{i+1,i}] .
 \end{equation}
 where $f(X_{i+1})$ and $f(Z_{i+1,i})$ are the probability density functions for $X_{i+1}$ and $Z_{i+1,i}$ respectively.
 $\mathfrak{E}_z$ actually accumulates the square error for each given $X_{i+1}$ and is a global error, while $\mathfrak{E}_x$ is a local error. \\
 \textbf{Theorem 1.} Under Assumptions 1 and 2, the optimal formations of $M_z$ and $M_x$ to minimize $\mathfrak{E}_z$ and $\mathfrak{E}_x$ are as follows:
 \begin{equation}
   \label{MzForm}
   \begin{matrix}
    M_z(Z_{i+1,i})=\mu_{i+1}+\Sigma_{i+1,i}{\Sigma_{i,i}}^{-1}(Z_{i+1,i}-\mu_{i}),
   \end{matrix}
 \end{equation}
 \begin{equation}
   \label{MxForm}
   M_x(\tilde{X}_{i+1,i})=\tilde{\mu}_{i+1}+\tilde{\Sigma}_{i+1,i}{\tilde{\Sigma}_{i,i}}^{-1}(\tilde{X}_{i+1,i}-\tilde{\mu}_{i}),
 \end{equation}
 where $\mu_{i+1}$ and $\mu_{i}$ stand for the expectation of $Z_{i+1,i+1}$ and $Z_{i+1,i}$, $\tilde{\mu}_{i+1}$ and $\tilde{\mu}_{i}$ for the expectation of $\tilde{X}_{i+1,i+1}$ and $\tilde{X}_{i+1,i}$, $\Sigma_{i+1,i}$ for the correlation matrix of $Z_{i+1,i+1}$ and $Z_{i+1,i}$, $\Sigma_{i,i}$ for the correlation matrix of $Z_{i+1,i}$ and $Z_{i+1,i}$, $\tilde{\Sigma}_{i+1,i}$ for the correlation matrix of $\tilde{X}_{i+1,i+1}$ and $\tilde{X}_{i+1,i}$, $\tilde{\Sigma}_{i,i}$ for the correlation matrix of $\tilde{X}_{i+1,i}$ and $\tilde{X}_{i+1,i}$. All of these symbols are trainable parameters of $M_z$ and $M_x$. 

 Its is given in the Appendix \ref{Theo1Proof}.

As Theorem 1 shows, the linear formations can achieve the least adjusting error and require little retraining overhead.

 \subsection{Principles of retraining loss function design}
 \label{lossFuncSec}
 Considering that the formations of $M_z$ and $M_x$ are so simple, we can make the retraining problem convex and its gradient Lipschitz continuous by sophisticatedly designing the loss function. There are two benefits to formulate the retraining problem as a convex and gradient-Lipschitz continuous one: preventing overfitting (i.e. there is no suboptimal point) and guaranteeing fast convergence with rate $O(\frac{1}{k})$, where $k$ is the number of iterations \cite{boyd2004convex, shapiro1996convergence}.

 Thus, we explore the requirements that the loss function should satisfy to ensure the convexity and gradient-Lipschitz continuousness and find that the convexity of the retraining process is not related to the model structure, but only to the design of a loss function. Since the trainable parameters in retraining stage are only involved in $M_z$ and $M_x$, we formulate the loss function in Definition 1, where the $\mathcal{L}_x(a,b)$ and $\mathcal{L}_z(a,b)$ are functions evaluating the distance between $a$ and $b$, and $\mathcal{P}_x$ and $\mathcal{P}_z$ stand for trainable parameters in $M_x$ and $M_z$ respectively
\\
 \textbf{Definition 1.}  The loss function is defined as follows: $\mathcal{L}(\mathcal{P}_x,\mathcal{P}_z)=\mathcal{L}_x(M_x(\tilde{X}_{i+1,i};\mathcal{P}_x),X_{i+1})+\mathcal{L}_z(M_z(Z_{i+1,i};\mathcal{P}_z),Z_{i+1,i+1})$. \\
 \textbf{Theorem 2.} If $\mathcal{L}_x(M_x(\tilde{X}_{i+1,i};\mathcal{P}_x),X_{i+1})$ and $\mathcal{L}_z(M_z(Z_{i+1,i};\mathcal{P}_z),Z_{i+1,i+1})$ are convex and gradient-Lipschitz continuous for $M_x(\tilde{X}_{i+1,i};\mathcal{P}_x)$ and $M_z(Z_{i+1,i};\mathcal{P}_z)$ respectively, $\mathcal{L}(\mathcal{P}_x,\mathcal{P}_z)$ is convex for $\mathcal{P}_x$ and $\mathcal{P}_z$, and its gradient is Lipschitz continuous.\\
Its proof is given in Appendix \ref{proofT2}.
 
 According to Theorem 2, the convexity of the loss function is only concerned with the convexity of $\mathcal{L}_x$ and $\mathcal{L}_z$ for its parameters, without concerning the structures of $Encoder$ and $Decoder$. Thus, it is easy to find proper formations of $\mathcal{L}_x$ and $\mathcal{L}_z$ to ensure convexity.
 According to Occam's razor principle \cite{walsh1979occam}, LARA chooses one of the simplest formations that satisfy the requirements in Theorem 2, i.e.,
 \begin{equation}
   \label{lossFunc}
     \mathcal{L}_x=(X_{i+1}-M_x(\tilde{X}_{i+1,i}))^2, \ 
     \mathcal{L}_z=(Z_{i+1,i+1}-M_z(Z_{i+1,i}))^2.
 \end{equation}
 
 \subsection{Limitation}
 The ruminate block helps LARA refresh the general knowledge learned by the old model, but may degrade its accuracy when the new distribution is very different from the old one.
 Section. \ref{DistImp} conducts the experiments and discuss the related issues.
  \begin{table*}[]
    \centering
    \setlength\tabcolsep{3.5pt}
    \renewcommand\arraystretch{1}
    \caption{\label{performance}Average precision, recall and the best F1 score results of LARA and baselines. `$^\ddagger$' indicates outdated models trained on old distribution data only, while `$^\dagger$' indicates the outdated models further retrained with small new distribution data. }
    \vspace{-4mm}
    \begin{tabular}{l|ccc|ccc|ccc|lll}
      \hline
                                                       & \multicolumn{3}{c|}{SMD}                                                           & \multicolumn{3}{c|}{J-D1}                                                          & \multicolumn{3}{c|}{J-D2}                                                                         & \multicolumn{3}{c}{SMAP}                                                    \\ \cline{2-13} 
                                                             & Prec                      & Rec                       & F1                         & Prec                      & Rec                       & F1                         & Prec                               & Rec                             & F1                         & \multicolumn{1}{c}{Prec} & \multicolumn{1}{c}{Rec} & \multicolumn{1}{c}{F1} \\ \hline
      Donut$^\ddagger$                                       & \multicolumn{1}{l}{0.793} & \multicolumn{1}{l}{0.811} & \multicolumn{1}{l|}{0.782} & \multicolumn{1}{l}{0.806} & \multicolumn{1}{l}{0.729} & \multicolumn{1}{l|}{0.734} & \multicolumn{1}{l}{0.919}          & \multicolumn{1}{l}{0.898}       & \multicolumn{1}{l|}{0.905} & 0.356                    & \textbf{1.000}          & 0.432                  \\

      Anomaly Transformer $^\ddagger$                               & \multicolumn{1}{l}{0.304} & \multicolumn{1}{l}{0.654} & \multicolumn{1}{l|}{0.415} & \multicolumn{1}{l}{0.331} & \multicolumn{1}{l}{0.852} & \multicolumn{1}{l|}{0.471} & \multicolumn{1}{l}{0.842}          & \multicolumn{1}{l}{ {0.986}} & \multicolumn{1}{l|}{0.907} & 0.297                & \textbf{1.000}          & 0.456                  \\
      OmiAnomaly$^\ddagger$                                  & 0.760                     & 0.778                     & 0.740                      & 0.847                     & 0.834                     & 0.815                      & 0.911                              & 0.898                           & 0.901                      & 0.809                    & \textbf{1.000}          & 0.869                  \\
      DVGCRN$^\ddagger$                                      & 0.578                     & 0.562                     & 0.530                      & 0.152                     & 0.569                     & 0.213                      & 0.333                              & 0.867                           & 0.420                      & 0.480                    & \textbf{1.000}          & 0.571                  \\
      ProS$^\ddagger$                                        & 0.344                     & 0.613                     & 0.407                      & 0.363                     & 0.818                     & 0.429                      & 0.678                              & 0.929                           & 0.781                      & 0.333                    & 0.992                   & 0.428                  \\
      VAE$^\ddagger$                                         & 0.576                     & 0.602                     & 0.575                      & 0.312                     & 0.716                     & 0.382                      & 0.716                              & 0.807                           & 0.738                      & 0.376                    & 0.992                   & 0.459                  \\
      MSCRED$^\ddagger$                                      & 0.508                     & 0.643                     & 0.484                      & 0.735                     & 0.859                     & 0.756                      & 0.894                              & 0.926                           & 0.909                      &  {0.820}                 & \textbf{1.000}          &  {0.890}            \\ 
      PUAD$^\ddagger$                                        & 0.923                     & \underline{0.993}         & \textbf{0.957}             & \underline{0.977}         & 0.790                     & 0.874                      & \textbf{0.990}                     & 0.775                           & 0.870                      & 0.908                    & \textbf{1.000}          & 0.952               \\
      TranAD$^\ddagger$                                      & 0.703                     & 0.594                     & 0.530	                    & 0.318                     & 0.851                     & 0.405                      & 0.729                              & 0.952                           & 0.797                      & 0.526                    & 0.838                   & 0.561               \\
      LARA-LD$^\ddagger$ & \multicolumn{1}{l}{0.613} & \multicolumn{1}{l}{0.885} & \multicolumn{1}{l|}{0.697} & \multicolumn{1}{l}{0.815} & \multicolumn{1}{l}{0.650} & \multicolumn{1}{l|}{0.682} & \multicolumn{1}{l}{0.828}          & \multicolumn{1}{l}{0.969}       & \multicolumn{1}{l|}{0.891} & 0.400                    & \textbf{1.000}          & 0.493                  \\
      LARA-LO$^\ddagger$ & 0.833                     & 0.665                     & 0.719                      & 0.876                     & 0.795                     & 0.793                      & 0.915                              & 0.955                           & 0.932                      & 0.733                    & 0.995                   & 0.802                  \\
      LARA-LV$^\ddagger$ & 0.704                     & 0.816                     & 0.741                      & 0.981                     & 0.825                     & 0.889	                     & 0.939                              & 0.861                           & 0.889                      & 0.812                    & \textbf{1.000}          & 0.846                  \\
      \hline
      Donut$^\dagger$                                        & \multicolumn{1}{l}{0.742} & \multicolumn{1}{l}{0.795} & \multicolumn{1}{l|}{0.764} & \multicolumn{1}{l}{0.950} & \multicolumn{1}{l}{0.650} & \multicolumn{1}{l|}{0.727} & \multicolumn{1}{l}{0.906}          & \multicolumn{1}{l}{0.913}       & \multicolumn{1}{l|}{0.904} & 0.502                    & \textbf{1.000}          & 0.578                  \\
      Anomaly Transformer$^\dagger$                          & \multicolumn{1}{l}{0.297} & \multicolumn{1}{l}{0.644} & \multicolumn{1}{l|}{0.407} & \multicolumn{1}{l}{0.324} & \multicolumn{1}{l}{0.852} & \multicolumn{1}{l|}{0.462} & \multicolumn{1}{l}{0.847}          & \multicolumn{1}{l}{ {0.986}}    & \multicolumn{1}{l|}{0.910}    & 0.295                    & \textbf{1.000}       & 0.453                  \\
      OmiAnomaly$^\dagger$                                   & 0.769                     & 0.887                     & 0.814                      & 0.827                     & 0.834                     & 0.800                      &  {0.945}                           & \underline{0.973}               &  {0.958}                   & 0.714                    & 0.995                   & 0.781                  \\
      DVGCRN$^\dagger$                                       & 0.573                     & 0.562                     & 0.521                      & 0.103                     & 0.790                     & 0.166                      & 0.311                              & 0.775                           & 0.371                      & 0.360                    & \textbf{1.000}          & 0.437                  \\
      ProS$^\dagger$                                         & 0.504                     & 0.533                     & 0.415                      & 0.375                     & 0.732                     & 0.373                      & 0.758                              & 0.803                           & 0.769                      & 0.574                    & 0.992                   & 0.620                  \\
      VAE$^\dagger$                                          & 0.482                     & 0.614                     & 0.488                      & 0.420                     & 0.732                     & 0.441                      & 0.686                              & 0.823                           & 0.711                      & 0.252                    & 0.992                   & 0.351                  \\
      MSCRED$^\dagger$                                       & 0.313                     & 0.796                     & 0.378                      & 0.969                     & 0.859                     & 0.895                      & 0.942                              & 0.926                           & 0.933                      & 0.793                    & \textbf{1.000}          & 0.857                  \\
      PUAD$^\dagger$                                         & 0.909                     & \textbf{0.995}            & 0.950                      & \textbf{0.978}            & 0.790                     & 0.874                      & \underline{0.983}                  & 0.677                           & 0.802                      & \underline{0.923}        & \textbf{1.000}          & \underline{0.960}      \\
      TranAD$^\dagger$                                       & 0.819                     & 0.834                     & 0.799                      & 0.767                     & 0.891                     & 0.781                      & 0.793                              & 0.964                           & 0.851                      & 0.726                    & \textbf{1.000}          & 0.819                  \\
      LARA-LD$^\dagger$  & \underline{0.925}         & \multicolumn{1}{l}{0.902} & \multicolumn{1}{l|}{0.913} & \multicolumn{1}{l}{0.878} & \multicolumn{1}{l}{0.928} & \multicolumn{1}{l|}{0.893} & \multicolumn{1}{l}{0.952}          & \multicolumn{1}{l}{0.924}       & \multicolumn{1}{l|}{0.936} & 0.788                    & \textbf{1.000}          & 0.863                  \\
      LARA-LO$^\dagger$  &  {0.921}                  & 0.952                     & 0.934                      &  {0.931}                  & \underline{0.969}         & \textbf{0.947}             & 0.942                              & \textbf{0.988}                  & \underline{0.964}          & 0.908                    & 0.995                   & 0.944                  \\ 
      LARA-LV$^\dagger$  & \textbf{0.945}            & 0.958                     & \underline{0.952}          & 0.914                     & \textbf{0.972}            & \underline{0.939}          & 0.976                              & 0.965                           & \textbf{0.970}             & \textbf{0.957}           & \textbf{1.000}          & \textbf{0.977}         \\
      \hline
      \end{tabular}%
  \end{table*}

  \begin{table*}[]
    \centering
    \setlength\tabcolsep{3.5pt}
    \renewcommand\arraystretch{1}
    \caption{\label{performance2}Compared few-shot LARA with baselines trained with the whole new-distribution dataset.}
    \vspace{-3mm}
    \begin{tabular}{l|ccc|ccc|ccc|lll}
      \hline
                                                            & \multicolumn{3}{c|}{SMD}                                                                 & \multicolumn{3}{c|}{J-D1}                                                                & \multicolumn{3}{c|}{J-D2}                                                                         & \multicolumn{3}{c}{SMAP}                                                    \\ \cline{2-13} 
                                                            & Prec                      & Rec                       & F1                               & Prec                      & Rec                             & F1                         & Prec                            & Rec                                & F1                         & \multicolumn{1}{c}{Prec} & \multicolumn{1}{c}{Rec} & \multicolumn{1}{c}{F1} \\ \hline
      JumpStarter                                           & \underline{0.943}         & 0.889                     & 0.907                            & 0.903                     & 0.927                           &  {0.912}                   & 0.914                           & 0.941                              & 0.921                      & 0.471                    & \underline{0.995}       & 0.526                  \\
      Donut                                                 & \multicolumn{1}{l}{0.809} & \multicolumn{1}{l}{0.819} & \multicolumn{1}{l|}{0.814}       & \multicolumn{1}{l}{0.883} & \multicolumn{1}{l}{0.628}       & \multicolumn{1}{l|}{0.716} & \multicolumn{1}{l}{0.937}       & \multicolumn{1}{l}{0.910}          & \multicolumn{1}{l|}{0.921} & 0.837                    & 0.859                   & 0.848                  \\
      Anomaly Transformer                                   & \multicolumn{1}{l}{0.894} & \multicolumn{1}{l}{0.955} & \multicolumn{1}{l|}{ {0.923}}    & \multicolumn{1}{l}{0.524} & \multicolumn{1}{l}{ {0.938}}    & \multicolumn{1}{l|}{0.673} & \multicolumn{1}{l}{0.838}       & \multicolumn{1}{l}{\textbf{1.000}} & \multicolumn{1}{l|}{0.910} & \underline{0.941}        & 0.994                   & \underline{0.967}      \\
      omniAnomaly                                           & 0.765                     & 0.893                     & 0.818                            &  {0.914}                  & 0.834                           & 0.855                      & 0.918                           & \underline{0.982}                  & 0.947                      & 0.736                    & 0.995                   & 0.800                  \\
      DVCGRN                                                & 0.482                     & 0.611                     & 0.454                            & 0.214                     & 0.599                           & 0.256                      & 0.412                           & 0.867                              & 0.444                      & 0.410                    & \textbf{1.000}          & 0.478                  \\
      ProS                                                  & 0.495                     & 0.623                     & 0.418                            & 0.210                     & 0.760                           & 0.306                      & 0.566                           & 0.886                              & 0.688                      & 0.287                    & 0.992                   & 0.395                  \\
      VAE                                                   & 0.541                     & 0.728                     & 0.590                            & 0.353                     & 0.550                           & 0.392                      & 0.686                           & 0.823                              & 0.711                      & 0.416                    & 0.992                   & 0.473                  \\
      MSCRED                                                & 0.813                     & 0.955                     & 0.874                            & 0.890                     & 0.859                           & 0.850                      & 0.956                           & 0.926                              &  {0.940}                   & 0.865                    & 0.991                   & 0.916                  \\ 
      PUAD                                                  & 0.920                     & \underline{0.992}         & \underline{0.955}                & \textbf{0.972}            & 0.896                           & 0.932                      & \textbf{0.990}                  & 0.782                              & 0.874                      & 0.925                    & \textbf{1.000}          & 0.961                  \\
      TranAD                                                & 0.926                     & \textbf{0.997}            & \textbf{0.961}                   & 0.841                     & 0.897                           & 0.833                      & 0.754                           & 0.965                              & 0.817                      & 0.804                    & \textbf{1.000}          & 0.892                  \\
      LARA-LD$^\dagger$ & \multicolumn{1}{l}{0.925} & \multicolumn{1}{l}{0.902} & \multicolumn{1}{l|}{0.913}       & \multicolumn{1}{l}{0.878} & \multicolumn{1}{l}{0.928}       & \multicolumn{1}{l|}{0.893} & \multicolumn{1}{l}{ {0.952}}    & \multicolumn{1}{l}{0.924}          & \multicolumn{1}{l|}{0.936} & 0.788                    & \textbf{1.000}          & 0.863                  \\
      LARA-LO$^\dagger$ &  {0.921}                  &  {0.952}                  & 0.934                            & \underline{0.931}         & \underline{0.969}               & \textbf{0.947}             & 0.942                           &  {0.988}                           & \underline{0.964}          &  {0.908}                 & \underline{0.995}       &  {0.944}            \\ 
      LARA-LV$^\dagger$ & \textbf{0.945}            & 0.958                     & 0.952                            & 0.914                     & \textbf{0.972}                  & \underline{0.939}          & \underline{0.976}               &  0.965                             & \textbf{0.970}             & \textbf{0.957}           & \textbf{1.000}          & \textbf{0.977}         \\
      \hline  
    \end{tabular}\vspace{0mm}
    \end{table*}

 \section{Experiments and their result analysis}
 Extensive experiments made on four real-world datasets demonstrate the following conclusions:
 \begin{enumerate}[itemindent=1em, listparindent=2em, leftmargin=0em]
  \item [1)] LARA trained by a few samples can achieve the highest F1 score compared with the SOTA methods and is competitive against the SOTA models trained with the whole subset (Sec.  \ref{PreAcc}).
  \item [2)] Both $M_z$ and $M_x$ improve the performance of LARA. Besides, the linear formations are better than other nonlinear formations (Sec.  \ref{sec:abla}), which is consistent with the mathematical analysis.
  \item [3)] Both the time and memory overhead of LARA are low (Sec.  \ref{sec:overhead}).
  \item [4)] LARA is hyperparameter-insensitive (Sec.  \ref{sec:hypersense}).
  \item [5)] The experimental results are consistent with the mathematical analysis of the convergence rate (Sec.  \ref{sec:convRate}). 
  \item [6)] LARA achieves stable performance increase when the amount of retraining data varies from small to large, while other methods' performances suddenly dip down due to overfit and only rebound with a large amount of retraining data (Sec.  \ref{retrAmt}).
  \item [7)] LARA significantly improves the anomaly detection performance for all the distribution-shift distances explored in Sec.  \ref{DistImp}.
 \end{enumerate}
 
 \subsection{Experiment setup} 
 \textbf{Baseline methods.} LARA is a generic framework that can be applied to cost-effectively retrain various existing VAEs, since the VAE-based methods exhibit a common property that they all learn two conditional distributions $p(\tilde{x}|z)$ and $p(z|x)$, where $x$, $\tilde{x}$ and $z$ denote the data sample, reconstructed sample and latent vector respectively. 
 In our experiments, LARA is applied to enable three state-of-the-art (SOTA) VAE-based detectors, Donut \cite{xu2018unsupervised}, OmniAnomaly \cite{su2019robust} and VQRAE \cite{kieu2022anomaly}, denoted by LARA-LD, LARA-LO and LARA-LV respectively. We compare LARA with transfer-learning-based methods, i.e., ProS \cite{kumagai2019transfer}, PUAD \cite{li2023prototype} and TranAD \cite{tuli2022tranad}, a signal-processing-based method called Jumpstarter \cite{ma2021jump}, a classical deep learning method called VAE \cite{DBLP:journals/corr/KingmaW13} and the SOTA VAE methods, i.e.,  Donut \cite{xu2018unsupervised}, OmniAnomaly \cite{su2019robust},  Multi-Scale Convolutional Recurrent Encoder-Decoder (MSCRED) \cite{zhang2019deep}, AnomalyTransformer \cite{DBLP:conf/iclr/XuWWL22} and Deep Variational Graph Convolutional Recurrent Network (DVGCRN) \cite{chen2022deep}. For more details of these baselines, please refer to the Appendix \ref{app:baselines}.
 

 \textbf{Datasets.} We use one cloud server monitoring dataset SMD \cite{su2019robust} and two web service monitoring datasets J-D1 and J-D2 \cite{ma2021jump}. Moreover, to verify the generalization performance, we use one of the most widely recognized anomaly detection benchmark, Soil Moisture Active Passive (SMAP) dataset \cite{hundman2018detecting}. For more details, please refer to appendix \ref{sec:dataset}.
 %

 \textbf{Evaluation metrics.} We use the widely used metrics for anomaly detection: precision, recall, and the best F1 score \cite{ma2021jump}. 
 \subsection{Prediction accuracy}
 \label{PreAcc}
 All of the datasets used in experiments consist of multiple subsets, which stand for different cloud servers for the web (SMD), different web services (J-D1, J-D2), and different detecting channels (SMAP). Different subsets have different distributions. Thus, the data distribution shift is imitated by fusing the data from different subsets. When verifying the model performances on shifting data distribution, the models are trained on one subset while they are retrained and tested on another one.
 When using a small amount of retraining data, the models are retrained by 1\% of data in a subset. When using enough retraining data, the models are retrained by the whole subset. For each method, we show the performance without retraining, retraining with few samples in Tab.  \ref{performance}, where the best performances are bold and the second best performances are underlined. Moreover, we compare the performance of few-shot LARA with the baselines trained with the whole new distribution dataset, which is shown in Tab.  \ref{performance2}. We obtain precision, recall and F1 score for best F1 score of each subset and compute the average metrics of all subsets. The "Prec" and "Rec" in Tab.  \ref{performance} stand for precision and recall respectively. For baseline method $\mathcal{A}$, $\mathcal{A}\ddagger$ denotes that $\mathcal{A}$ is trained on the old distribution and tested on the new distribution without retraining. $\mathcal{A}\dagger$ denotes that $\mathcal{A}$ is trained on the old distribution and tested on the new distribution with retraining via a small amount of data from the new distribution and $\mathcal{A}$ denotes that the model is trained on new distribution with enough and much data and tested on new distribution. As JumpStarter is a signal-processing-based method of sampling and reconstruction, retraining is not applicable to JumpStarter. Besides, transfer between old distribution and new one is not applicable to JumpStarter. Thus, Jumpstarter is not shown in Tab. \ref{performance}. Retraining with small-amount data means using only 1\% data of each subset from new distribution: 434 time slots of data for SMD, 80 time slots of data for J-D1, 74 time slots of data for J-D2, 43 time slots of data for SMAP.

\begin{table*}[]
  \centering
  \caption{\label{AblationStudy}Ablation study results. The best results are in bold.}
    \vspace{-3mm}
  \begin{tabular}{l|ccc|ccc|ccc|ccc}
    \hline
    \multirow{2}{*}{} & \multicolumn{3}{c|}{SMD}                         & \multicolumn{3}{c|}{J-D1}                        & \multicolumn{3}{c|}{J-D2}                        & \multicolumn{3}{c}{SMAP}                         \\ \cline{2-13} 
                      & Prec           & Rec            & F1             & Prec           & Rec            & F1             & Prec           & Rec            & F1             & Prec           & Rec            & F1             \\ \hline
    remove $M_z$      & 0.855          & \textbf{0.973} & 0.892          &  {0.913}    & 0.861          &  {0.879}    & 0.842          & 0.975          & 0.901          &  {0.751}    & \textbf{1.000} &  {0.833}    \\
    remove $M_x$      & 0.764          & 0.829          & 0.786          & 0.812          & 0.826          & 0.818          & 0.858          & 0.976          & 0.904          & 0.632          & \textbf{1.000} & 0.694          \\
    replace with MLP  &  {0.917}    & 0.933          &  {0.925}    & 0.466          &  {0.929}    & 0.530          & 0.848          & \textbf{0.990} & 0.904          & 0.600          & \textbf{1.000} & 0.674          \\
    replace with SA   & 0.875          & 0.733          & 0.797          & 0.477          & 0.884          & 0.519          &  {0.863}    & 0.978          &  {0.905}    & 0.602          & \textbf{1.000} & 0.685          \\
    LARA-LO              & \textbf{0.921} &  {0.952}    & \textbf{0.934} & \textbf{0.931} & \textbf{0.969} & \textbf{0.947} & \textbf{0.942} &  {0.988}    & \textbf{0.964} & \textbf{0.978} &  {0.990}    & \textbf{0.983} \\ \hline
    \end{tabular}%
    \vspace{-3mm}
\end{table*}

 As Tab. \ref{performance} shows, LARA$^\dagger$ achieves the best F1 score on all of the datasets, when compared with baselines retrained with a small amount of data. Moreover, LARA$^\dagger$ also achieves competitive F1 scores when compared with the baselines trained by the whole subset of new distribution, as shown in Tab. \ref{performance2}.

Besides, LARA exhibits strong anti-overfitting characteristics. When retrained by small-amount data, LARA$^\dagger$ dramatically improves the F1 score even with 43 time slots of data, while the F1 score of some baselines reduces dramatically after retraining with small-amount data due to overfitting. 
There is an interesting phenomenon shown in Tab. \ref{performance}. Some methods show better performance when transferred to new distribution without retraining than those trained and tested on new distribution. This verifies that there is some general knowledge among different distributions and it makes sense to utilize a part of an old model.

As LARA-LO$^\dagger$ achieves better performance than LARA-LD$^\dagger$, in the following, we look into LARA-LO$^\dagger$ and analyze its performance from different aspects.

\subsection{Ablation study}
\label{sec:abla}
We verify the effectiveness of $M_z$ and $M_x$ by separately eliminating them and comparing the resulting performance with LARA's. Moreover, to verify the optimality of the linear formation which is proven mathematically in Sec.  \ref{adjustFunc}, we substitute the linear layer with multi-layer perceptrons (MLP) and self attention (SA) and compare their performance with LARA's. The ablation study results are shown in Tab. \ref{AblationStudy}. It is clear that 1) removing either $M_z$ or $M_x$ can lead to large F1 score drop on all four datasets, and 2) the linear formation performs better than the non-linear formations, which is consistent with the mathematical analysis result in Sec.  \ref{adjustFunc}.

 \subsection{Time and memory overhead}
 \label{sec:overhead}
\begin{figure*}[t]
  \centering
  \subfigure[Time (s) and Memory (* 1000B) Overhead for Retraining]{
      \includegraphics[width=0.265\linewidth]{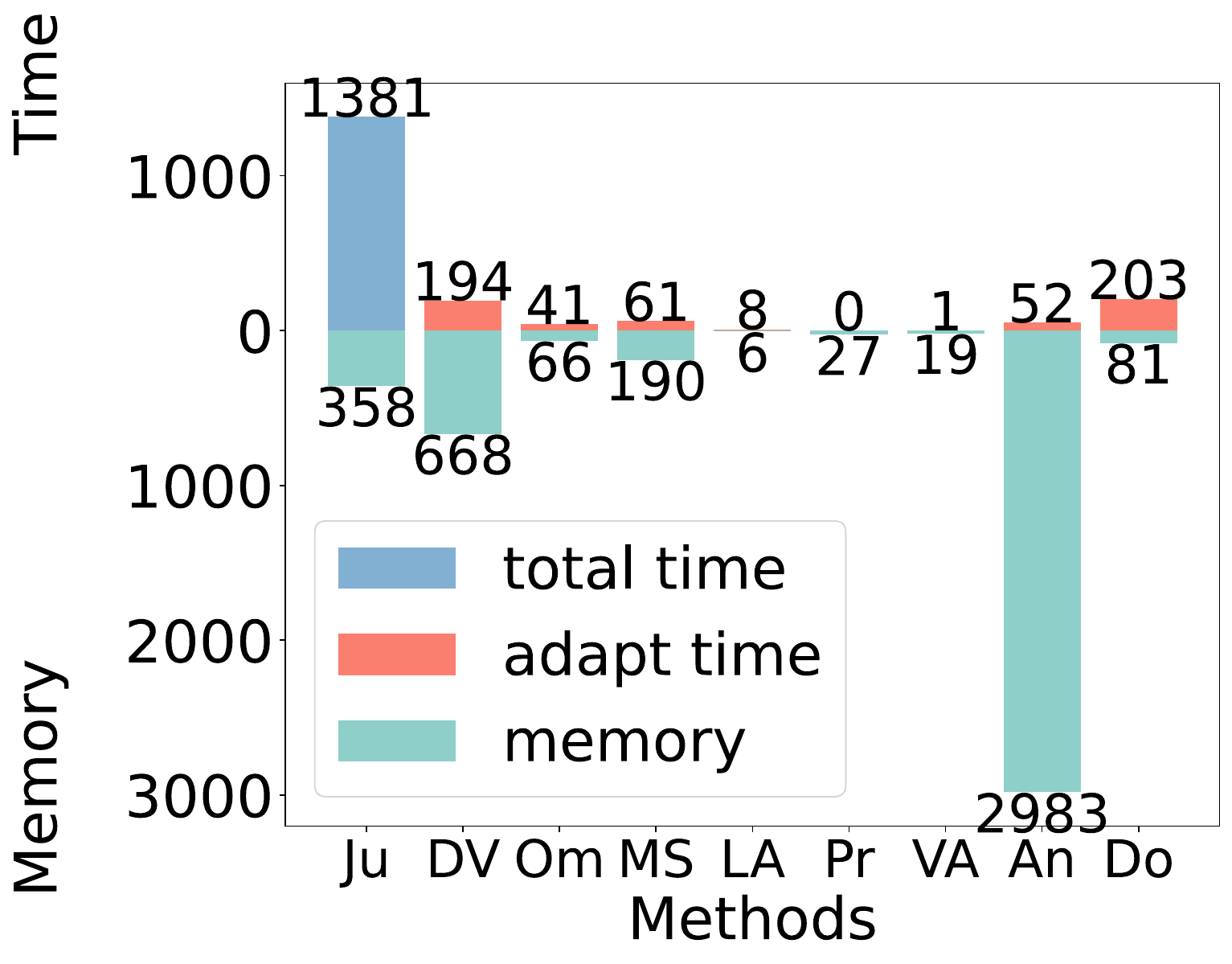}
      \label{fig:time}
  }
  \hfill
  \subfigure[Ratios of Retraining Time (s) and Memory (* 1000B) Overhead to Training Ones]{
      \includegraphics[width=0.265\linewidth]{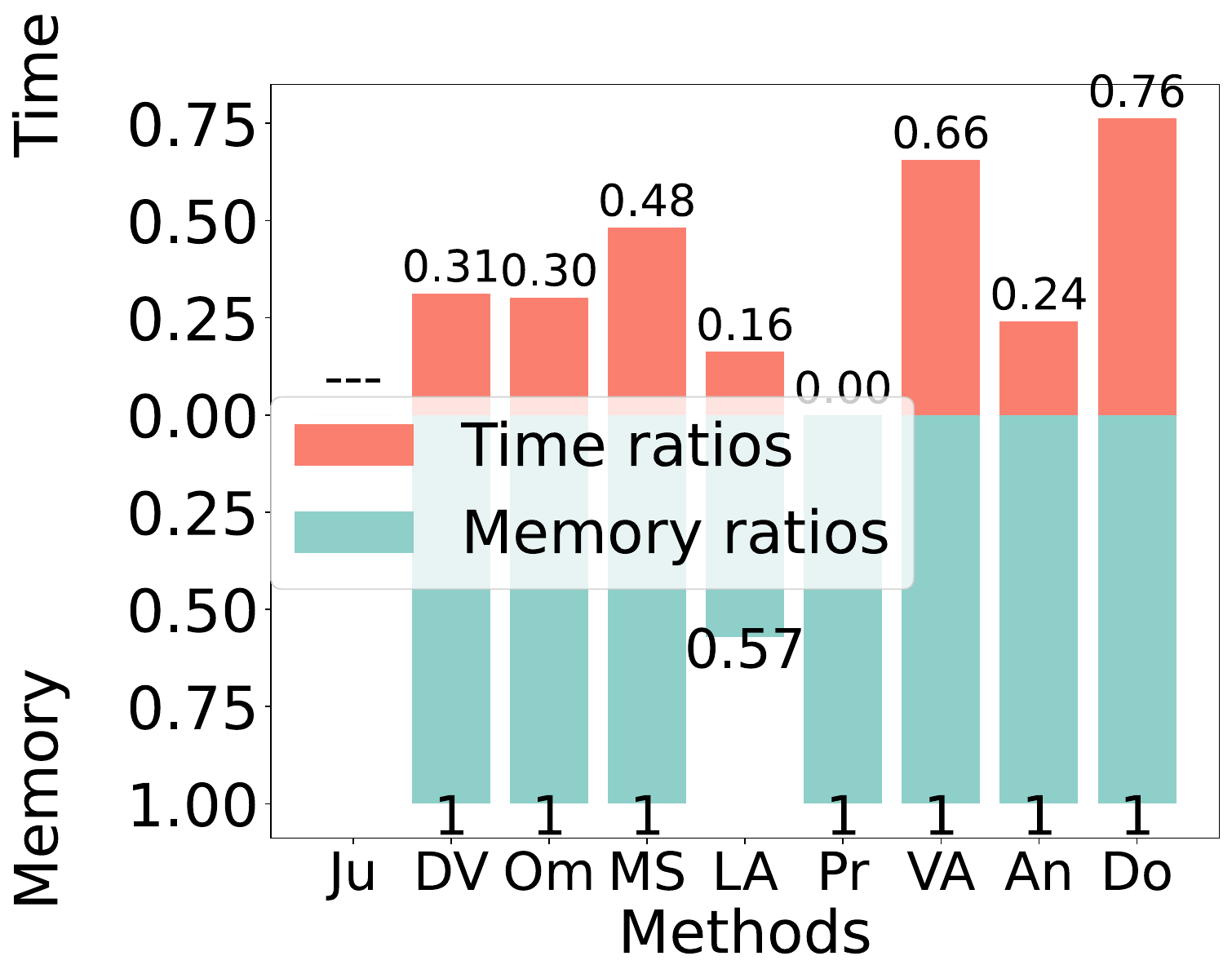}
      \label{fig:timeRatio}
  }
  \hfill
  \subfigure[The F1 score for different combination of hyperparameters]{
  \includegraphics[width=0.32\linewidth]{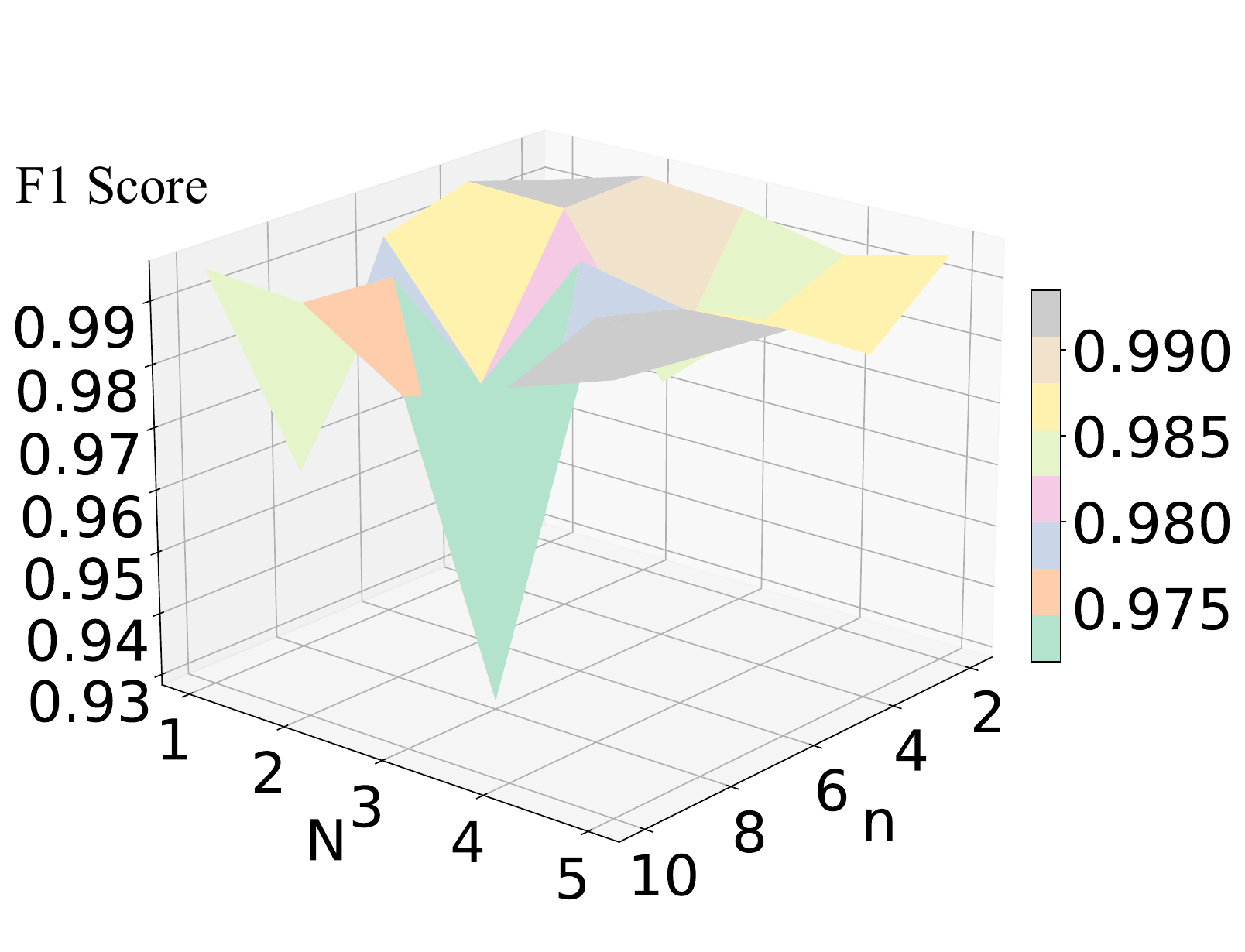}
  \label{fig:hyper}
  }
  \hfill
  \subfigure[Convergence rate of LARA]{
      \includegraphics[width=0.285\linewidth]{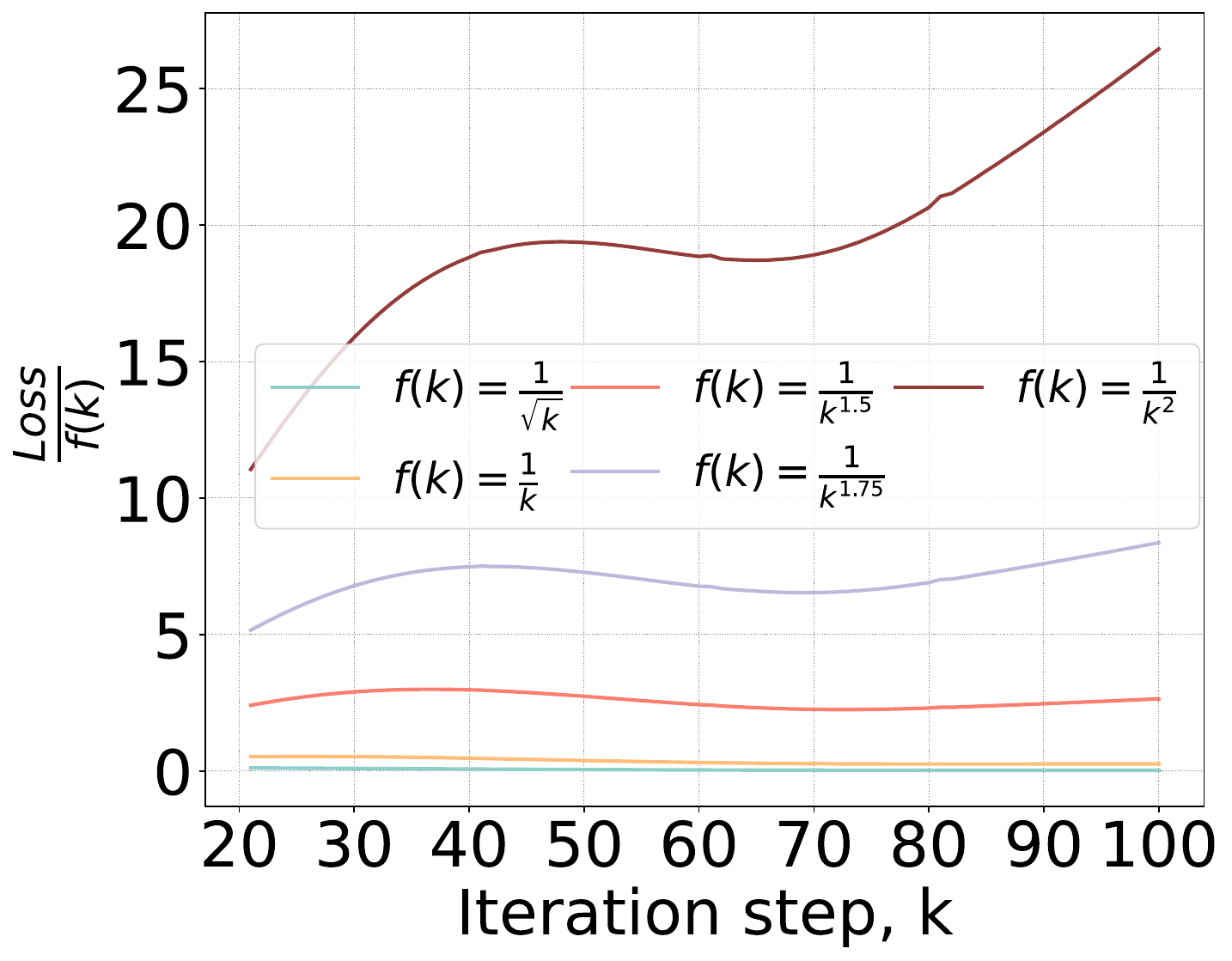}
      \label{fig:ConvRate}
  }
  \hfill
  \subfigure[The F1 score for different retraining data amount]{
      \includegraphics[width=0.28\linewidth]{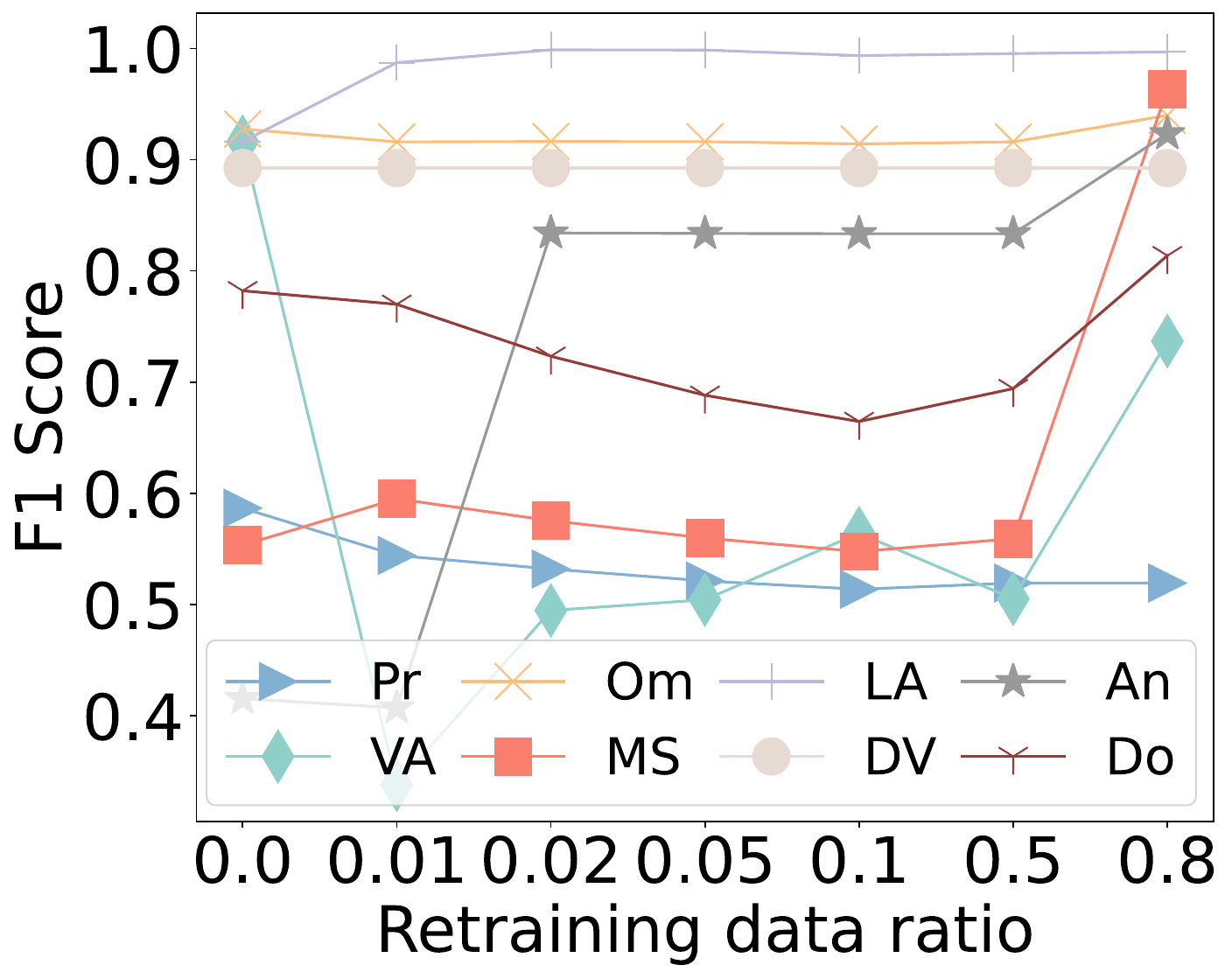}
      \label{fig:adapt}
  }
  \hfill
  \subfigure[The F1 score for different distribution distance]{
      \includegraphics[width=0.3\linewidth]{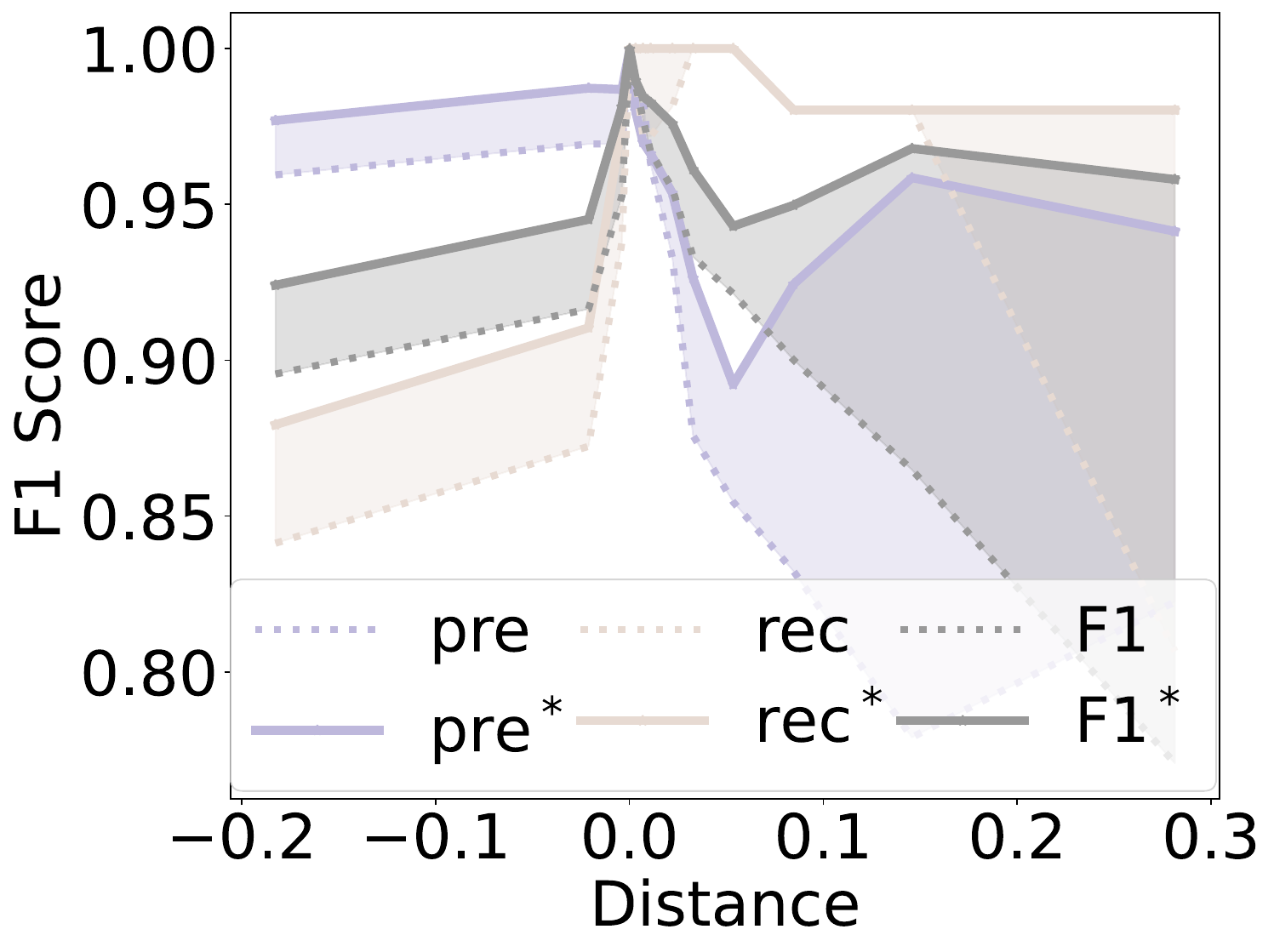}
      \label{fig:distance}
  }
  \hfill
  \vspace{-4mm}
  \caption{Due to space constraints, we use the first two letters as the shorthand for each method. (a) As the memory overhead of JumpStarter, AnomalyTransformer, and MSCRED are dramatically larger than the others, to show the memory overhead of other methods clearly, we divide their memory overhead by 10. (b) The ratios of retraining memory and time overhead to training memory and time overhead. (d) The ratio of loss to iteration count varies with the number of iteration count. (e) The x-label is the proportion of retraining data in new distribution data. (f) In the legend, we use pre, rec, F1 to denote precision, recall and F1 score before retraining and use pre$^*$, rec$^*$, F1$^*$ to denote them after retraining. }
\end{figure*}

 We use an Intel(R) Xeon(R) CPU E5-2620 @ 2.10GHz CPU and a K80 GPU to test the time overhead. For the neural network methods, we use the profile tool to test the memory overhead for retraining parts of a model. As for JumpStarter, which is not a neural network, we use htop to collect its maximized memory consumption. We show the result in Fig.\ref{fig:time}. From it, we can conclude that LARA achieves the least retraining memory consumption and relatively small time overhead. There are only two methods whose time overhead is lower than LARA: VAE and ProS. However, the F1 score of VAE is low. When the distance between source and target domains is large, the F1 score of ProS is low. 
 Moreover, it is also interesting to look into the ratios of retraining time and memory overhead to the training ones, which is shown in Fig.\ref{fig:timeRatio}. Since Jumpstarter is not a neural network and the retraining process is not applicable to Jumpstarter, the ratio of Jumpstarter is omitted.
 As Fig.\ref{fig:timeRatio} shows, LARA achieves the smallest retraining memory ratios and the second smallest retraining time ratios. 

 \subsection{Hyperparameter sensitivity}
 \label{sec:hypersense}
 There are two important hyperparameters in LARA: $n$, the number of restored historical data samples for each data sample from new distribution, and $N$, the number of samples when calculating statistical mean. We test the F1 score of LARA by setting $n$ and $N$ as the Cartesian product of $n$ from 1 to 5 and $N$ from 2 to 10. We show the result in Fig.\ref{fig:hyper}. Within the search space, the maximum F1 score is only 0.06 higher than the minimum F1 score. With occasional dipping down, the F1 score basically remains at a high level for different hyperparameter combinations, thus verifying LARA's insensitivity to hyperparameters by utilizing Taguchi’s experimental design method and intelligent optimization methods \cite{gao2018dendritic,chen2021hierarchical,luo2021generalized}.
 
 \subsection{Convergence rate}
 \label{sec:convRate}
To verify the convergence rate of LARA, which is analyzed theoretically in Sec.  \ref{lossFuncSec}, the loss for each iteration step is divided by $f(k)=\frac{1}{\sqrt{k}}$, $\frac{1}{k}$, $\frac{1}{k^{1.5}}$, $\frac{1}{k^{1.75}}$ and $\frac{1}{k^{2}}$ respectively, which is shown in Fig.\ref{fig:ConvRate}. $k$ denotes the iteration count.
 As shown in Fig.\ref{fig:ConvRate}, when the loss is divided by $\frac{1}{\sqrt{k}}$, $\frac{1}{k}$ and $\frac{1}{k^{1.5}}$, the quotients remain stable as the iterations grow. Thus, the convergence rate is $O(\frac{1}{k})$, which is consistent with mathematical analysis result.

\subsection{Performance after multiple retrainings}
To verify that LARA can maintain good and stable performance after multi-time retrainings, we first train LARA on one subset and retrain it on eight subsets consecutively. As verified in Tab. ~\ref{Tab:retrain}, LARA can maintain good and stable performances for the first 8 iterations, though there are some moderate performance fluctuations.
\begin{table}[]
  \centering
  \caption{\label{Tab:retrain}The performance of LARA after retraining for several times.} \vspace{-2mm}
  \begin{tabular}{l|cccc}
  \hline
  Retraining Times & 2     & 4     & 6     & 8     \\ \hline
  F1 Score         & 0.979 & 0.939 & 0.985 & 0.925 \\ \hline
  \end{tabular}%
  \vspace{-3mm}
\end{table}

 \subsection{Impact of retraining data amount on LARA's performance}
 \label{retrAmt}
 We use a subset in SMD to explore the impact of retraining data amount on the performance of LARA$\dagger$. We first use data from one subset to train all of the models. After that, we use different ratios of data from another subset to retrain the models and test them on the testing data in this subset. We show the result in Fig.\ref{fig:adapt}. When the retraining ratio is 0, there is no retraining. As the figure shows, LARA significantly improves the F1 score even with 1\% data from new distribution, while the F1 score of other methods dramatically dips down due to overfitting. After the first growth of the F1 score of LARA, its F1 score remains stable and high, while the F1 score of many other methods only grows after using enough data.

 \subsection{Impact of transfer distance between old and new distributions}
 \label{DistImp}
 When training a model on dataset $\mathcal{A}$ and testing it on dataset $\mathcal{B}$, the closer the distributions of $\mathcal{A}$ and $\mathcal{B}$ are, the higher the model's accuracy is.
  Inspired by this insight, we have defined a directed transfer distance to quantify the distance to transfer a specific model $\mathcal{C}$ from dataset $\mathcal{A}$ to dataset $\mathcal{B}$ (trained on $\mathcal{A}$ and tested on $\mathcal{B}$) as $Distance_{\mathcal{A}\rightarrow\mathcal{B}}(\mathcal{C})$. This can be quantified by comparing the F1 score degradation when testing on $\mathcal{B}$ after training on $\mathcal{A}$, with testing and training both on $\mathcal{B}$. Let $F1\ score^*$ denote the F1 score for both training and testing on training and testing set of $\mathcal{B}$ and $F1\ score$ denote the F1 score for training on $\mathcal{A}$ and testing on test set of $\mathcal{B}$ without any retraining. Intuitively, the $F1\ score^*$ represents the performance that model $\mathcal{C}$ should have achieved on Dataset $\mathcal{B}$ and provides a benchmark for the measurement. We compare the $F1\ score$ with the benchmark and get our transfer distance: $Distance_{\mathcal{A}\rightarrow\mathcal{B}}(\mathcal{C})=\frac{F1\ score^* - F1\ score}{F1\ score}$.
     Different from traditional distance definition, this distance is asymmetrical. In other words, $Distance_{\mathcal{A}\rightarrow\mathcal{B}}(\mathcal{C})$ is different from $Distance_{\mathcal{B}\rightarrow\mathcal{A}}(\mathcal{C})$. Besides, it can be negative: when the data quality of dataset $\mathcal{A}$ is better than that of dataset $\mathcal{B}$'s training set, and their distributions are very similar, the transfer distance $Distance_{\mathcal{A}\rightarrow\mathcal{B}}(\mathcal{C})$ can be negative.
 Then we show how the transfer distance between old and new distributions impacts the F1 score of LARA in Fig.\ref{fig:distance}. Intuitively, when the distance is zero, LARA works best and there is little accuracy difference before and after retraining. Besides, the larger the distance is, the greater the accuracy difference before and after retraining is. 
As Fig.\ref{fig:distance} shows, LARA can significantly improve the anomaly detection performance in the range of transfer distance explored in the experiments. 
 
 \section{Conclusion}
 In this paper, we focus on the problem of web-service anomaly detection when the normal pattern is highly dynamic, the newly observed retraining data is insufficient and the retraining overhead is high.
 To solve the problem, we propose LARA, which is light and anti-overfitting when retraining with little data from the new distribution. We innovatively formulate the model retraining problem as a convex problem and solves it with a ruminate block and two light adjusting functions. The convexity prevents overfitting and guarantees fast convergence, which also contributes to the small retraining overhead. The ruminate block makes better use of historical data without storing them. The adjusting functions are mathematically and experimentally proven to achieve the least adjusting errors.
Extensive experiments conducted on four real-world datasets demonstrate the anti-overfitting and small-overhead properties of LARA. It is shown that LARA retrained with a tiny amount of new data is competitive against the state-of-the-art models trained with sufficient data. This work represents a significant advance in the area of web-service anomaly detection.

\begin{acks}
  This work was supported by the Key Research Project of Zhejiang Province under Grant 2022C01145, the National Science Foundation of China under Grants 62125206 and U20A20173, and in part by Alibaba Group through Alibaba Research Intern Program.
\end{acks}

\bibliographystyle{ACM-Reference-Format}
\bibliography{sample-base}

\appendix

\section{Implementation details}
All of methods retraining with small-amount data on SMD dataset use 434 data samples. All of methods retraining with small-amount data on J-D1 dataset use 80 data samples. All of methods retraining with small-amount data on J-D2 dataset use 74 data samples. All of methods retraining with small-amount data on SMAP dataset use 43 data samples.
We mainly use grid search to tune our hyperparameters. The searching range for $n$ is from 1 to 5. The searching range for $N$ is from 1 to 10. The searching range of learning rate is {0.001,0.002,0.005,0.008,0.01}. The search range for batch size is {50,100,400}. The searching range of input window length is {40,50,80,100}. The searching range of hidden layer is {1,2,3,5}.

\textbf{Hyperparameters.} The hyperparameters are listed in Tab. \ref{hyper}, where $N$ stands for the number of samples when estimating the expectation in Eq.\ref{expectationMonte} and $n$ stands for the number of restored historical data for each newly-observed data sample.

\section{Proof of Theorem 1}
\label{Theo1Proof}
\emph{Proof of Theorem 1.} 
In the following, we take $M_z$ as an example to prove the formation in Eq.(\ref{condDis}) is optimal. Then, the formation of $M_x$ can be inferred in a similar way but given $Z_{i+1,i}$ in each step. We firstly use the following lemmas 1 and 2 to show that the optimal formation of $M_z(Z_{i+1,i})$ is $\mathbb{E}(Z_{i+1,i+1}|Z_{i+1,i})$. Then, if Assumption 2 holds, we can substitute the $\mathbb{E}(Z_{i+1,i+1}|Z_{i+1,i})$ with the Gaussian conditional expectation and then get the Eq.(\ref{condDis}).

\begin{equation}
    \label{condDis}
    \begin{matrix}
     M_z(Z_{i+1,i})=\mu_{i+1}+\Sigma_{i+1,i}{\Sigma_{i,i}}^{-1}(Z_{i+1,i}-\mu_{i})
    \end{matrix}
  \end{equation}

\textbf{Lemma 1.} The $\mathfrak{E}_z$ can be further transformed into $\mathbb{E}[\mathcal{A}^2]+\mathbb{E}[\mathcal{B}^2]$, where $\mathcal{A}$ and $\mathcal{B}$ are $M_z(Z_{i+1,i})-\mathbb{E}(Z_{i+1,i+1}|Z_{i+1,i})$ and $\mathbb{E}(Z_{i+1,i+1}|Z_{i+1,i})-Z_{i+1,i+1}$ respectively.

\emph{Proof of Lemma 1.} According to the definition of expectation, the mapping error can be transformed into $\mathbb{E}_x(\mathbb{E}_{Z_{i+1,i},Z_{i+1,i+1}}((M_z(Z_{i+1,i})-Z_{i+1,i+1})^2|x))$. For clarity, we use the subscript of $\mathbb{E}$ to denote the variable for this expectation. Furthermore, the two-layer nested expectations can be reduced to the form of a unified expectation $\mathbb{E}_{Z_{i+1,i},Z_{i+1,i+1}}(M_z(Z_{i+1,i})-Z_{i+1,i+1})^2$. We plus and minus $E(Z_{i+1,i+1}|Z_{i+1,i})$ at the same time and then we get $\mathbb{E}_{Z_{i+1,i},Z_{i+1,i+1}}(\mathcal{A}-\mathcal{B})^2$. We expand the $\mathbb{E}_{Z_{i+1,i},Z_{i+1,i+1}}(\mathcal{A}-\mathcal{B})^2$ and then get $\mathbb{E}[\mathcal{A}^2]-2\mathbb{E}[(\mathcal{A})(\mathcal{B})]+\mathbb{E}[\mathcal{B}^2]$. We take a further look at the middle term $\mathbb{E}((\mathcal{A})(\mathcal{B}))$ and find it is equal to zero, as shown in the next paragraph. Thus, the Lemma 1 holds.

Now we prove that the $\mathbb{E}[(\mathcal{A})(\mathcal{B})]$ is equal to 0. Since there are two variables in $\mathbb{E}[(\mathcal{A})(\mathcal{B})]$ : the $Z_{i+1,i+1}$ and the $Z_{i+1,i}$, we can transform the $\mathbb{E}[(\mathcal{A})(\mathcal{B})]$ into 
$\mathbb{E}_{Z_{i+1,i}}[\mathbb{E}_{Z_{i+1,i+1}}[(\mathcal{A})(\mathcal{B})|Z_{i+1,i}]]$. When $Z_{i+1,i}$ is given, the first multiplier in the inner expectation is a constant and can be moved to the outside of the inner expectation. Then we get $\mathbb{E}[(\mathcal{A})(\mathcal{B})] = \mathbb{E}_{Z_{i+1,i}}[\mathcal{A}\cdot\mathbb{E}_{Z_{i+1,i+1}}[\mathcal{B}|Z_{i+1,i}]]$. Recalling $\mathcal{B}=\mathbb{E}(Z_{i+1,i+1}|Z_{i+1,i})-Z_{i+1,i+1}$, when $Z_{i+1,i}$ is given, $\mathbb{E}(Z_{i+1,i+1}|Z_{i+1,i})$ is a constant and can be moved to the outside of the inner expectation. Then we get $\mathbb{E}[(\mathcal{A})(\mathcal{B})] = \mathbb{E}_{Z_{i+1,i}}[\mathcal{A} \cdot (\mathbb{E}[Z_{i+1,i+1}|Z_{i+1,i}]-\mathbb{E}[Z_{i+1,i+1}|Z_{i+1,i}])]$. Thus, $\mathbb{E}[(\mathcal{A})(\mathcal{B})]$ is equal to $0$.

\textbf{Lemma 2.} $M_z(Z_{i+1,i})=\mathbb{E}(Z_{i+1,i+1}|Z_{i+1,i})$ is the optimal solution to minimize $\mathfrak{E}_z$.

\emph{Proof of Lemma 2.} According to Lemma 1, $\mathfrak{E}_z=\mathbb{E}[(M_z(Z_{i+1,i})-\mathbb{E}(Z_{i+1,i+1}|Z_{i+1,i}))^2]-\mathbb{E}[\mathcal{B}^2]$. Only the first term involves $M_z$. Since the first term is greater than or equal to 0, when $M_z(Z_{i+1,i})$ takes the formation of $\mathbb{E}(Z_{i+1,i+1}|Z_{i+1,i})$, $\mathbb{E}[(M_z(Z_{i+1,i})-\mathbb{E}(Z_{i+1,i+1}|Z_{i+1,i}))^2]$ reaches its minimum value of 0.

\begin{table}[]
  \centering
  \caption{\label{hyper}The default hyperparmeter values for LARA.}
  \begin{tabular}{c|c|c|c}
    \hline
    \multicolumn{1}{c|}{\textbf{Hyperparameter}} & \textbf{Value} & \multicolumn{1}{c|}{\textbf{Hyperparameter}} & \textbf{Value} \\ \hline
    Batch size                                   & 100            & $n$                                            & 3              \\ \hline
    Learning rate                                & 0.001          & $N$                                            & 10             \\ \hline
    \end{tabular}%
  \end{table}

  \begin{table*}[ht]
    \centering
    \caption{\label{multi-seed}Average precision, recall and F1 score for 10 seeds on SMD. }
    \begin{tabular}{l|l|l|l|l|l|l|l|l|l}
      \hline
                & LARA  & MSCRED & Omni  & ProS  & VAE   & JumpStarter & DVGCRN & AnomalyTrans & Donut \\ \hline
      precision & 0.882 & 0.859  & 0.791 & 0.246 & 0.200 & -           & 0.762  & 0.849        & 0.770 \\ \hline
      recall    & 0.971 & 0.967  & 0.904 & 1.000 & 0.754 & -           & 0.698  & 0.962        & 0.894 \\ \hline
      F1        & 0.923 & 0.908  & 0.837 & 0.334 & 0.298 & -           & 0.709  & 0.901        & 0.819 \\ \hline
      \end{tabular}%
    \end{table*}
    \begin{table}[]
      \centering
      \caption{\label{tab:distance} The distance between training and testing set}
      \begin{tabular}{l|llll}
        \hline
      \textbf{}                & \textbf{SMD} & \textbf{J-D1} & \textbf{J-D2} & \textbf{SMAP} \\
      \hline
      Expectation of KL        & 0.109        & 0.466         & 0.436         & 0.582         \\
      Standard deviation of KL & 0.024        & 0.079         & 0.026         & 0.230        \\
      \hline
      \end{tabular}%
  \end{table}
\section{Proof of Theorem 2}
\label{proofT2}
\emph{Proof sketch of Theorem 2.}
We take a further look at the formation of $M_z$ and $M_x$. They can be transformed into affine functions of $\mathcal{P}_z$ and $\mathcal{P}_x$. According to Stephen Boyd \cite{boyd2004convex}, when the inner function of a composite function is an affine function and the outer function is a convex function, the composite function is a convex function. Thus, if $\mathcal{L}_x$ and $\mathcal{L}_z$ are convex functions, $\mathcal{L}(\mathcal{P}_x,\mathcal{P}_z)$ is convex. Moreover, as the affine function is gradient Lipschitz continuous, if the $\mathcal{L}_x$ and $\mathcal{L}_z$ are gradient Lipschitz continuous, $\mathcal{L}(\mathcal{P}_x,\mathcal{P}_z)$ is gradient Lipschitz continuous by using the chain rule for deviation. 

\section{Proof of Ruminate block}
\label{proofTh3}
\emph{Proof of Ruminate block.} 
Since the $\bar{X}_i $ is reconstructed from $X_{i+1} $, it is assumed that they have the approximately same latent vector. It is also assumed that the reconstructed data can approximate the original data.
The proof is shown in Eq.\ref{expectationproof}-Eq.\ref{varianceproof}, where $p(z)$ follows the normal distribution which is also assumed by \cite{DBLP:journals/corr/KingmaW13}.
\begin{equation}
  \label{expectationproof}
  \begin{matrix}
    \mathbb{E}(\tilde{Z}_{i+1} |X_{i+1} )&=\int_z \frac{p(\tilde{X}_{i+1,i} |z) p(\bar{X}_{i}|Z_{i+1,i}) p(z) z}{\int_z p(\tilde{X}_{i+1,i} |z) p(\bar{X}_{i}|Z_{i+1,i}) p(z) dz } dz\\
 &=  \frac{\mathbb{E}_{z\sim p(z)}[p(\tilde{X}_{i+1} |z)p(\bar{X}_i|Z_{i+1,i})z]}{\mathbb{E}_{z\sim p(z)}[p(\tilde{X}_{i+1} |z)p(\bar{X}_i|Z_{i+1,i})]}
  \end{matrix}
\end{equation}
\begin{equation}
  \label{varianceproof}
  \begin{matrix}
    \mathbb{E}(\tilde{Z}_{i+1} ^T\tilde{Z}_{i+1} |X_{i+1} )&=\int_z \frac{p(\tilde{X}_{i+1,i} |z) p(\bar{X}_{i}|Z_{i+1,i}) p(z)z^Tz}{\int_z p(\tilde{X}_{i+1,i} |z) p(\bar{X}_{i}|Z_{i+1,i}) p(z) dz } dz\\
    &= \frac{\mathbb{E}_{z\sim p(z)}[p(\tilde{X}_{i+1,i} |z) p(\bar{X}_{i}|Z_{i+1,i}) z^Tz]}{\mathbb{E}_{z\sim p(z)}[p(\tilde{X}_{i+1,i} |z) p(\bar{X}_{i}|Z_{i+1,i})]}\\
  \end{matrix}
\end{equation}

\section{Multi-run experiments for different seeds}
 Random seeds introduce significant uncertainty in the training of neural networks. To verify that the results in Tab. \ref{performance} are not obtained occasionally, we make multi-run experiments for 10 randomly chosen seeds on SMD. The average precisions, recalls and F1 scores are shown in Tab. \ref{multi-seed}. Since Jumpstarter is not a neural network, the multi-seed experiments are not applicable to it. It is proven that the average metrics of multi-run experiments are similar to the results in Tab. \ref{performance}.

\section{Introduction of the baselines}
\label{app:baselines}
\begin{itemize}[itemindent=1em, listparindent=2em, leftmargin=0em]
  \item \emph{Donut \cite{xu2018unsupervised}} is one of the prominent time series anomaly detection methods, which improves the F1 score by using modified ELBO, missing data injection and MCMC imputation. 
  \item \emph{OmniAnomaly \cite{su2019robust}} is one of widely-recognized anomaly detection methods. It uses a recurrent neural network to learn the normal pattern of multivariate time series unsupervised. 
  \item \emph{MSCRED \cite{zhang2019deep}} is another widely-recognized anomaly detection method, which uses the ConvLSTM and convolution layer to encode and decode a signature matrix and can detect anomalies with different lasting length.
  \item  \emph{DVGCRN  \cite{chen2022deep}} is a recent method and has reported high F1 score on their datasets. It models channel dependency and stochasticity by an embedding-guided probabilistic generative network. Furthermore, it combines Variational Graph Convolutional Recurrent Network (VGCRN) to model both temporal and spatial dependency and extend the VGCRN to a deep network.
  \item \emph{ProS  \cite{kumagai2019transfer}} aims at improving the anomaly detection performance on target domains by transferring knowledge from related domains to deal with target one. It is suitable for semi-supervised learning and unsupervised one. For fairness, we use its unsupervised learning as the others are unsupervised.
  \item \emph{JumpStarter  \cite{ma2021jump}} also aims to shorten the initialization time when distribution changes. It mainly uses a Compressed Sensing technique. Moreover, it introduces a shape-based clustering algorithm and an outlier-resistant sampling algorithm to support multivariate time series anomaly detection.
  \item \emph{AnomalyTransformer \cite{DBLP:conf/iclr/XuWWL22}} is one of the latest and strongest anomaly detection methods. It proposes an Anomaly-Attention mechanism and achieves high F1 scores on many datasets.
  \item \emph{VAE  \cite{DBLP:journals/corr/KingmaW13}} is one of the classic methods for anomaly detection and is the root of lots of nowadays outstanding methods. It uses stochastic variational inference and a learning algorithm that scales to large datasets.
  \item \emph{Variants of LARA. } LARA is used to retrain two deep VAE-based methods: Donut \cite{xu2018unsupervised} and OmniAnomaly \cite{su2019robust}, which are denoted by LARA-LD and LARA-LO respectively.
  To verify the improvement of our retraining approach, we design a controlled experiment. We use LARA-{X}$^\ddagger$ to denote that LARA-{X} is trained on old distribution and tested on new distribution without retraining, where {X} can be LD or LO. We use LARA-{X}$^\dagger$ to denote that LARA-{X} is trained on old distribution and tested on new distribution with retraining with a small amount of data.
\end{itemize}

\section{Introduction of datasets}
\label{sec:dataset}
We introduce each dataset in the following. Moreover, for each dataset, we compute the kernel density of each subset. Subsequently, we compute the KL divergence of the density estimations between every training-testing pair of subsets to measure the distance between the training and testing sets. As shown in Tab. ~\ref{tab:distance}, the distances between the training and testing sets for each dataset are large and variable.
\begin{itemize}[itemindent=1em, listparindent=2em, leftmargin=0em]
  \item \emph{Server Machine Dataset (SMD) \cite{su2019robust}} is a 5-week-long dataset. It is collected from a large Internet company. This dataset contains 3 groups of entities. SMD has the data from 28 different machines, forming 28 subsets. In this dataset, the data distributions of the training data and retraining data are the most different. 
  \item \emph{Datasets provided by JumpStarter (J-D1 and J-D2) \cite{ma2021jump}} are collected from a top-tier global content platform. The datasets consist of the monitoring metrics of its 60 different services, forming 60 subsets. In this dataset, the data distributions of the training data and retraining data are the most similar. 
  \item \emph{Soil Moisture Active Passive (SMAP) \cite{hundman2018detecting}} consists of real spacecraft telemetry data and anomalies from the Soil Moisture Active Passive satellite. 
\end{itemize}

\section{Anomaly detection mechanism}
To detect anomalies, LARA computes the reconstruction error of each sample in a time series. Subsequently, it uses POT \cite{siffer2017anomaly} to determine a threshold of reconstruction error. When the reconstruction error is greater than the threshold, the sample is inferred as an anomaly.

\section{Related work}
Anomaly detection is to find the outlier of a distribution~\cite{pang2021deep,zhang2023self,yang2023dcdetector}. We first overview popular anomaly detection approaches for static normal patterns. 
We then summarize transfer learning, few-shot learning, statistical learning and signal-processing-based methods, which can be used to solve a normal pattern changing problem. There may be a concern that online learning is also a counterpart of LARA. However, the peak traffic of web services is extremely high, and online learning is inefficient and struggles to deal with it in real time.

\textbf{Anomaly detection for static normal pattern.} The current popular anomaly detection methods can be divided into classifier-based \cite{gao2020robusttad,DBLP:conf/iclr/HendrycksG17, DBLP:conf/nips/LeeLLS18, DBLP:conf/iclr/GrathwohlWJD0S20,DBLP:conf/icml/RuffGDSVBMK18, DBLP:journals/corr/abs-2006-00339,shen2020timeseries} and reconstructed-based ones \cite{li2022learning,su2019robust,zhang2019deep,shen2021time,tian2019learning} roughly. Though these methods achieve high F1 scores for a static normal pattern, their performance decays as the normal pattern changes.

\textbf{Transfer learning for anomaly detection.} One solution for the normal pattern changing problem is transfer learning \cite{DBLP:conf/iccv/MotiianPAD17, DBLP:conf/icml/HoffmanTPZISED18, long2016unsupervised, saito2018maximum, long2018conditional,Cao_2023_ICCV}. These methods transfer knowledge from source domains to a target domain, which enables a high accuracy with few data in the target domain. However, these methods do not work well when the distance between source and target domain is large. Moreover, transfer learning mainly transfers knowledge of different observing objects and there is no chronological order of these observations, while the different distributions in this work are observations at different times of the same observing object. The nearer historical distribution is usually more similar to the present one and contains more useful knowledge. But transfer learning ignores this aspect.

\textbf{Continuous learning for anomaly detection.} Among the works of anomaly detection equipped with continuous learning, there are also some models utilizing VAE to replay the historical data \cite{wiewel2019continual}. However, LARA distinguishes itself by different overhead, different real-time capabilities and different application scenarios. Firstly, LARA focuses on updating the model with lighter retraining overhead. Thus, LARA chooses one of the simplest formations to fine-tune the model and formulate the retraining process as a convex problem to accelerate model convergence speed and reduce computational overhead. Secondly, LARA aims to shorten the unavailable period of the model due to the outdated problem. Thus, LARA reduces the amount of data required for retraining to shorten the time for newly-observed data collection. Moreover, LARA chooses one of the simplest formations, which can be quickly fitted. In general, LARA can update a model in less than 8s. Thirdly, The retraining process of LARA only needs to be triggered in the situation where there is a distribution change, while continuous learning has to keep an eye on the newly collected data consistently no matter whether there is a distribution shift.

\textbf{Few-shot learning for anomaly detection.} Few-shot learning determines to extract general knowledge from different tasks and improve the performance on the target task with few data samples. Metaformer \cite{wu2021learning} is one of the recently prominent few-shot learning based anomaly detection methods, which uses MAML \cite{finn2017model} to find an ideal initialization. However, few-shot learning has a similar problem as transfer learning (i.e. it overlooks the chronological orders). Moreover, it needs to store lots of outdated historical-distributed data and costs lots of storage space.

\textbf{Statistics-based anomaly detection.} Traditional statistics-based methods \cite{zhang2021cloudrca,choffnes2010crowdsourcing, guha2016robust, pang2015lesinn} do not need training data and have light overhead, which are not bothered by a normal pattern changing problem. However, these methods rely on certain assumptions and are not robust in practice \cite{ma2021jump}.

\textbf{Signal-processing-based anomaly detection.} Fourier transform \cite{zhao2019automatic} can only capture global information, while wavelet analysis \cite{alarcon2001anomaly} can capture local patterns but is very time-consuming. PCA and Kalman Filtering \cite{ndong2011signal} are the most classical signal-processing techniques but are not competitive in detecting anomalies in variational time series. JumpStarter \cite{ma2021jump} is a recent SOTA method in this category, but it suffers from heavy inferring time overhead and can not deal with a heavy traffic load in real time.
\end{document}